\documentclass[twoside]{article}

\usepackage[accepted]{aistats2025}
\usepackage[round]{natbib}

\usepackage{hyperref}       
\usepackage{amsmath}
\usepackage{amsfonts}       
\usepackage{mathtools}

\usepackage{booktabs}
\usepackage{threeparttable}
\usepackage{makecell}

\usepackage{graphicx}
\usepackage{caption}
\usepackage{wrapfig}
\usepackage{subcaption}
\usepackage{float}  

\usepackage{tikz}
\usepackage{tikz-3dplot}
\usetikzlibrary{positioning}
\usepackage{xcolor}

\usepackage[title]{appendix}
\usepackage{titlesec}

\newcommand{\RR}{\mathbb{R}}
\newcommand{\R}[1]{\mathbb{R}^{#1}}
\newcommand{\mbf}[1]{\mathbf{#1}}
\newcommand{\mtc}[1]{\mathcal{#1}}
\newcommand{\bds}[1]{\boldsymbol{#1}}
\newcommand*{\dif}{\mathop{}\!\mathrm{d}}
\newcommand{\matern}{Matérn}
\newcommand{\addls}{\addlinespace[0.45em]}

\begin{document}

%
\runningtitle{Adaptive RKHS Fourier Features for Compositional GP Models}

%
\runningauthor{Xinxing Shi, Thomas Baldwin-McDonald, Mauricio A. Alvarez}

\twocolumn[

\aistatstitle{Adaptive RKHS Fourier Features for Compositional Gaussian Process Models}

\aistatsauthor{%
  Xinxing Shi\textsuperscript{1,2} \And
  Thomas Baldwin-McDonald\textsuperscript{1} \And
  Mauricio A. Álvarez\textsuperscript{1} 
}

\aistatsaddress{%
  \\ \textsuperscript{1} Department of Computer Science \\
  The University of Manchester, UK\\
  \textsuperscript{2}\texttt{xinxing.shi@postgrad.manchester.ac.uk}
} 

]

\begin{abstract}
Deep Gaussian Processes (DGPs) leverage a compositional structure to model non-stationary processes. DGPs typically rely on local inducing point approximations across intermediate GP layers. Recent advances in DGP inference have shown that incorporating global Fourier features from the Reproducing Kernel Hilbert Space (RKHS) can enhance the DGPs' capability to capture complex non-stationary patterns. 
This paper extends the use of these features to compositional GPs involving linear transformations. In particular, we introduce Ordinary Differential Equation(ODE)--based RKHS Fourier features that allow for adaptive amplitude and phase modulation through convolution operations. This convolutional formulation relates our work to recently proposed deep latent force models, a multi-layer structure designed for modelling nonlinear dynamical systems. By embedding these adjustable RKHS Fourier features within a doubly stochastic variational inference framework, our model exhibits improved predictive performance across various regression tasks.
\end{abstract}

\section{INTRODUCTION}
\label{sec:Intro}

Gaussian Processes (GPs) provide a principled Bayesian framework for function approximation, making them particularly useful in many applications requiring uncertainty calibration \citep{gpml}, such as Bayesian optimisation \citep{bo2012} and time-series analysis \citep{timeseries-gp}. Despite offering reasonable uncertainty estimation, shallow GPs often struggle to model complex, non-stationary processes present in practical applications. To overcome this limitation, Deep Gaussian Processes (DGPs) employ a compositional architecture by stacking multiple GP layers, thereby enhancing representational power while preserving the model's intrinsic capability to quantify uncertainty \citep{dgp2013}. However, the conventional variational formulation of DGPs heavily depends on local inducing point approximations across GP layers \citep{svgp2009, dgp2017}, which hinder the model from capturing the global structures commonly found in real-world scenarios.

Incorporating \emph{Fourier features} into GP models has shown promise in addressing this challenge in GP inference due to the periodic nature of these features. A line of research uses Random Fourier Features (RFFs) \citep{rahimi2007random} of stationary kernels to convert the original (deep) GPs into Bayesian networks in weight space \citep{lazaro2010sparsespectrum, gal2015improving, cutajar2017random}. Building on this concept within a sparse variational GP framework, recent advancements in inter-domain GPs \citep{inter-domain-gp2009, van2020framework} directly approximate the posterior of the original GPs by introducing \emph{fixed} Variational Fourier Features (VFFs) through process projection onto a Reproducing Kernel Hilbert Space (RKHS)\citep{vffs, iddgp2020}.

VFFs are derived by projecting GPs onto a different domain. The original GP posterior that these VFFs attempt to approximate remains within the same functional space as the original GP. In this setting, the VFFs produce a set of static basis functions determined by a fixed set of frequency values. To enhance these features and introduce greater flexibility, we propose a generalised approach that incorporates Fourier features into inter-domain GPs through linear transformations, such as convolution operations. 

In this paper, we focus on a type of GP characterised as the output of a convolution operation between a smoothing kernel and a latent GP. An example of this construction is the Latent Force Model (LFM) \citep{lfms}, in which the smoothing kernel corresponds to the Green's function associated with an Ordinary Differential Equation (ODE). By incorporating RKHS Fourier features into this framework,  we derive adaptive global features inspired by the ODE,  allowing for the optimisation of amplitudes and phases. We name the obtained features \emph{Variational Fourier Response Features} (VFRFs) since they are derived from the output of a linear system. To enhance the capability of our model, we further use these adaptive features in a compositional GP model that stacks multiple LFMs, also known as Deep LFM (DLFM) \citep{dlfm2021}. This hierarchical structure facilitates more precise and robust modelling of complex, non-stationary data. Our experimental results on both synthetic and real-world data demonstrate that incorporating these ODE-inspired RKHS Fourier features improves upon the standard practice of using VFFs. 

\section{BACKGROUND}

This section reviews concepts and preliminaries relevant to this work and establishes the notation used throughout the subsequent discussions.

\subsection{Sparse Variational Gaussian Process}

A GP $f(\cdot)\sim\mtc{GP}(m(\cdot),k(\cdot,\cdot'))$ places probability measures on a function space $\{f:\R{D} \rightarrow \RR\}$ \citep{gpml}. Its behaviour is characterised by the mean function $m:\R{D}\rightarrow\RR$ and the covariance function $k: \R{D}\times\R{D}\rightarrow\RR$. The evaluation of the function $f(\cdot)$ at an input of interest $\mbf{x}$ is a random variable denoted as $f(\mbf{x})\in\RR$. Given a dataset of inputs $\mbf{X} = [\mbf{x}_n]_{n=1}^N\in \R{N\times D}$ and the corresponding measurements $\mathbf{y}=[y_n]_{n=1}^N\in\R{N}$, we assume $\mbf{y}$ is observed from a noise-corrupted GP: $y_n = f(\mbf{x}_n) + \epsilon, \epsilon\sim\mtc{N}(\epsilon\mid 0, \varepsilon^2)$, where $\varepsilon^2$ is the noise variance. The exact inference for the posterior distribution $p(f\mid\mbf{y})$ suffers from $\mathcal{O}(N^3)$ time complexity and is limited to Gaussian likelihoods. 

Sparse Variational Gaussian Processes (SVGPs) \citep{svgp2009, gpbigdata2013,hensman2015scalable} provide a scalable inference framework by introducing a small set of $M (\ll N)$ inducing points $\mbf{Z}=[\mbf{z}_m]_{m}^M\in\R{M\times D}$ and the corresponding inducing variables $\mbf{u}=[u(\mbf{z}_m)]_{m=1}^M\in\R{M}$ from the GP prior, i.e., $p(\mbf{u})=\mtc{N}(\mbf{u}\mid \mbf{0},\mbf{K_{ZZ}})$. A variational distribution $q(\mbf{u})=\mtc{N}(\mbf{u}\mid \mbf{m}, \mbf{S})$ is employed to approximate the posterior process 
$q(f(\mbf{x}))=\int p(f(\mbf{x})\mid\mbf{u})q(\mbf{u})\dif\mbf{u}=\mtc{N}(f\mid\tilde{m},\tilde{\Sigma})$, where
\begin{equation}
\begin{aligned}
    \tilde{\mu}(\mbf{x})
    &=
    m(\mbf{x})+k_{\mbf{x}\mbf{Z}}\mbf{K}^{-1}_{\mathbf{ZZ}}\mbf{m},
    \\
    \tilde{\Sigma}(\mbf{x},\mbf{x}')
    &=
    k_{\mbf{x}\mbf{x}'}+\mbf{k}_{\mbf{x}\mbf{Z}}\mbf{K}^{-1}_{\mbf{ZZ}}(\mbf{S}-\mbf{K_{ZZ}})\mbf{K}^{-1}_{\mbf{ZZ}}\mbf{k}_{\mbf{Z}\mbf{x}'}.
\end{aligned}
\label{eq:svgp-post}
\end{equation}
SVGPs learn the optimal placement of the inducing points and the variational distribution by maximising an Evidence Lower BOund (ELBO) of $\log p(\mbf{y}\mid\mbf{X})$.

\subsection{Variational Fourier Features}

Inter-domain GPs \citep{lazaro2009inter, alvarez2010efficient, van2020framework} extend the domain of inducing variables by integrating the GP $f$ with a deterministic inducing function $g$:
\begin{equation}
    u(\mathbf{z}) = \int_{\mathbb{R}^D} g(\mathbf{x},\mathbf{z}) f(\mathbf{x})\dif\mathbf{x},\quad\mathbf{z}\in\mathbb{R}^{D'}.
    \label{general-interdomain-gp}
\end{equation}
This formulation allows for a redefinition of inducing variables $\mathbf{u}=u(\mbf{Z})$, which still share the GP prior, albeit with alternative expressions of $\mbf{k}_{\mbf{x}\mbf{Z}}$ and $\mbf{K_{ZZ}}$ used in \eqref{eq:svgp-post}. By choosing various functions for $g$, inter-domain GPs facilitate the construction of vector-valued basis functions $k(\cdot,\mbf{Z})$ for more informative feature extraction while maintaining the standard SVGP framework.

VFFs \citep{vffs} define each inter-domain inducing variable $u_m$ of $\mbf{u}$ by projecting the original GP $f$ onto a Fourier basis: $u_m= \langle \phi_m, f\rangle_{\mathcal{H}}$, where $\langle\cdot,\cdot\rangle_{\mathcal{H}}$ denotes the \matern\ RKHS inner product on an interval $[a,b]$, and $\phi_m$ is the $m$-th entry of a truncated Fourier basis 
\begin{equation}
\begin{aligned}
    \bds{\phi}(x)  = & 
    [1, \cos(z_1(x-a)),\cdots,\cos(z_M(x-a)),\\ & \sin(z_1(x -a)),\cdots,\sin(z_M(x-a))],
\end{aligned}
\label{eq:fourier-basis}
\end{equation}
with $x$ a scalar input. In this setting, $\mathbf{z}=[z_m]_{m=1}^M$ are $M$ \emph{inducing frequencies}, analogous to inducing points in the SVGP context. The projection of $f$ onto this basis results in sinusoidal terms in the cross-covariance, i.e.,  $\mathrm{Cov}[f(\cdot), u_m]= \phi_m(\cdot)$ due to the reproducing property of the RKHS.

\begin{figure*}[th]
    \centering
    \includegraphics[width=0.8\linewidth]{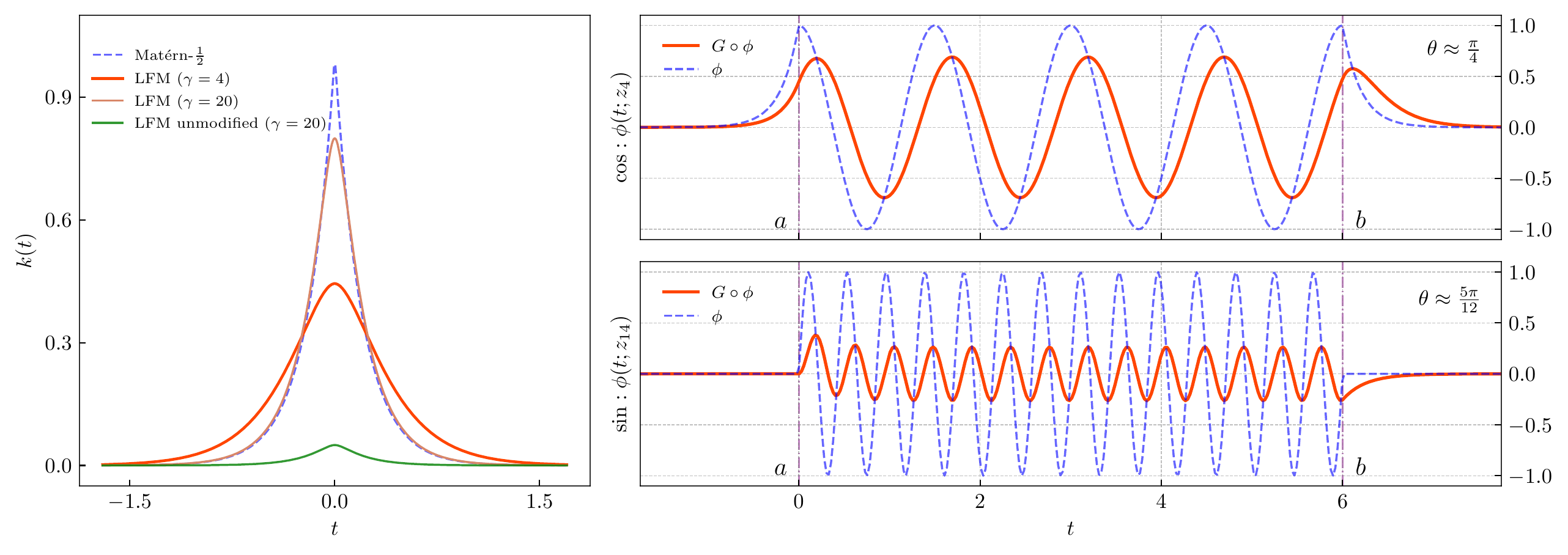}
    \caption{Covariance functions of LFMs (left) and Variational Fourier Response Features (VFRFs) (right). The latent force $u(t)$ uses a \matern-$\tfrac{1}{2}$ kernel with length-scale $l=0.2$ (left dashed). \textbf{Left}: The centred kernel of the input latent force (dashed) and the output process $f(t)$ of LFMs with different ODE parameters $\gamma$ (solid). Unlike the LFM kernel induced by \eqref{eq:lfm-equation} (green), the modified LFM kernel from \eqref{eq:ode-ab} can revert to the input \matern-$\tfrac{1}{2}$ kernel if increasing $\gamma$ (red to brown). \textbf{Right}: VFRFs ($G\circ\phi$, red solid) and VFFs ($\phi$, blue dashed) with different inducing frequencies: $z_m=\tfrac{8\pi}{b-a}$ (upper) and $\tfrac{28\pi}{b-a}$ (lower). The upper panel depicts the cosine basis with a phase delay $\theta\approx\tfrac{\pi}{4}$ to the VFF, while the lower panel displays the sine basis with a phase delay $\theta\approx\frac{5\pi}{12}$.}
    \label{fig:basis-functions}
\end{figure*}
\subsection{Latent Force Model}

An LFM \citep{lfms} is a GP model integrating differential equations to model dynamic physical systems probabilistically. Based on prior physical knowledge, a single-output LFM assumes the system's output $f(t)$ is influenced by $Q$ \emph{latent forces} $\{u_q(t)\}_{q=1}^Q$ through differential equations. Commonly, a first-order LFM uses the following form of ODE \citep{fastkernel}
\begin{equation}
    \frac{\dif f(t)}{\dif t} + \gamma f(t) = \sum_{q=1}^Q S_{q} u_q(t),
    \label{eq:lfm-equation}
\end{equation}
where $\gamma> 0$ is a decay parameter, and $S_{q}\in\RR$ is a sensitivity parameter. The solution for the output $f(t)$ takes the form of weighted convolution integrals
$f(t) = \sum_{q=1}^Q S_{q}\int_0^t G(t-\tau)u_q(\tau)\dif\tau$, where $G(\cdot)$ denotes the Green's function associated with the ODE.

Latent forces are presumed to follow GP priors, $u_q(t)\sim\mtc{GP}(0, k_q(t, t'))$, leading to a covariance function for the outputs $k_{f}(t,t') = \sum_{q=1}^Q S^2_{q}\iint G(t-\tau)k_q(\tau,\tau')G(t'-\tau')\dif\tau\dif\tau'.$ 
For some types of covariance functions $k_q(t,t')$, e.g., the radial basis function (RBF), $k_f(t,t')$ can either be computed explicitly \citep{lawrence2006modelling} or approximated by using convolved RFFs \citep{fastkernel,rahimi2007random}. By plugging the physics-informed kernels into the GP posterior, LFMs embed domain-specific knowledge into the learning process and can utilise the sparse approximation techniques in GP inference.

\section{METHODOLOGY}

This section describes integrating RKHS Fourier features into compositional GPs, with a specific focus on LFMs within the SVGP framework. We start by adapting the conventional ODE used in LFMs to incorporate VFFs as a special instance of our model (Section~\ref{sec:Modify-ODE}). Details on VFRFs are provided in Section~\ref{sec:VFRFs}. We then extend our model from a single-layer to a hierarchical structure in Section~\ref{sec:DLFM}.

\subsection{LFMs with Modified ODEs}
\label{sec:Modify-ODE}

In this work, we focus on a dynamical system modelled by a potential first-order ODE without loss of generality
\begin{equation}
    \beta \frac{\dif f(t)}{\dif t} + \alpha f(t) = u(t),
    \label{eq:ode-ab}
\end{equation}
where $\alpha, \beta$ are positive coefficients and $u(t)\sim\mtc{GP}(0, k(t,t'))$ represents an unknown latent force with a \matern\ kernel with half-integer order. The Green's function of \eqref{eq:ode-ab}
is $G(t)=\frac{1}{\beta}\exp(-\frac{\alpha}{\beta}t)=\frac{1}{\beta}\exp(-\gamma t)$. We introduce $\gamma=\frac{\alpha}{\beta}$ to remain consistent with the decay parameter in \eqref{eq:lfm-equation}. Unlike the conventional formulation \eqref{eq:lfm-equation}, which involves a weighted sum of multiple latent forces, the modified ODE \eqref{eq:ode-ab} simplifies it to a single process $u(t)$. It can be trivially decomposed into distinct latent forces if necessary. Moreover, we will further show that, by introducing coefficients $\alpha$ and $\beta$ in our model, the output process $f$ can revert to a GP with VFFs as $\beta\rightarrow0^{+}$. This formulation enables practitioners to apply the proposed approach in scenarios where prior knowledge of the system is limited and there is no prior knowledge indicating if the dynamics encoded in the kernel accurately reflect the observed data.

A solution $f(t)$ can be expressed as a convolution integral 
\begin{equation}
    f(t) = \int_{-\infty}^t G(t-\tau)u(\tau)\dif \tau=G\circ u,
    \label{eq:f(t)}
\end{equation}
where the integral's lower limit is extended to negative infinity to maintain variance near the origin, though it can be adjusted based on data range or prior knowledge in practice. The covariance function of the output process $f(\cdot)$ is derived by applying the convolution operator to the kernel $k$'s arguments, respectively:
\begin{equation}
\begin{aligned}
    & \text{Cov}[f(t), f(t')] = \\ &\quad \int_{-\infty}^{t'}\int_{-\infty}^t G(t-\tau)k(\tau,\tau')G(t'-\tau')\dif\tau\dif\tau'.
    \label{eq:gkg}
\end{aligned}
\end{equation}
The covariance function (also called LFM kernel in this paper) can be calculated analytically if $k$ is a Matérn kernel with half-integer orders. We give the closed-form covariances in Table~\ref{tab:gkg} of Appendix~\ref{app:lfm-kernel}. 

\begin{figure*}[ht]
\centering
\definecolor{myblue}{RGB}{0, 71, 171}
\definecolor{myorange}{RGB}{221, 0, 0}
\definecolor{myyellow}{RGB}{218, 165, 32}
\definecolor{mylightblue}{RGB}{169, 169, 239}
\definecolor{mypurple}{RGB}{220, 223, 240}
\begin{subfigure}{\textwidth}
\centering
\begin{tikzpicture}[scale=0.85,roundnode-1/.style={circle, fill=green!20, font=\small}, roundnode-2/.style={circle, fill=orange!20, font=\small},
squarednode/.style={rectangle, draw=black!50,font=\small},
]
    \node[roundnode-1] (x1) at (0,0) {$t_1$};
    \node[roundnode-1] (x2) at (0,-1) {$t_2$};
    \node[roundnode-2] (f11) at (3,0) {$f^1_1$};
    \node[roundnode-2] (f12) at (3,-1) {$f^1_2$};
    \draw[thick, ->] (x1) -- (f11);
    \draw[thick, ->] (x2) -- (f12);

    \node[squarednode] (sum-1) at (4.5, -0.5) {$\Sigma$};
    \draw[thick,->] (f11) -- (sum-1);
    \draw[thick,->] (f12) -- (sum-1);

    \node[roundnode-1] (g11) at (6, 0) {$g^1_1$};
    \node[roundnode-1] (g12) at (6, -1.) {$g^1_2$};
    \draw[thick,->] (sum-1) -- (g11);
    \draw[thick,->] (sum-1) -- (g12);

    \node[roundnode-2] (f21) at (9,0) {$f^2_1$};
    \node[roundnode-2] (f22) at (9,-1) {$f^2_2$};
    \draw[thick,->] (g11) -- (f21);
    \draw[thick,->] (g12) -- (f22);

    \node[squarednode] (sum-2) at (10.5, -0.5) {$\Sigma$};
    \draw[thick,->] (f21) -- (sum-2);
    \draw[thick,->] (f22) -- (sum-2);
    \node[roundnode-1] (g21) at (12,-0.5) {$g^2_1$};
    \draw[thick,->] (sum-2) -- (g21);
    \node[circle, fill=mypurple!50] (y) at (13.5,-0.5) {$y$};
    \draw[thick,->] (g21) -- (y);
\end{tikzpicture}
\caption{Inter-domain deep Gaussian process (IDDGP)}
\label{fig:subfig-idgp}
\end{subfigure}

\begin{subfigure}{\textwidth}
\centering
\vspace{1.1em}
\begin{tikzpicture}[scale=0.85, roundnode-1/.style={circle, fill=green!20, font=\small}, roundnode-2/.style={circle, fill=orange!20, font=\small}, 
squarednode/.style={rectangle, draw=black!50,font=\small}, squarednode-2/.style={rectangle, fill=red!20, draw=black!50,font=\small}
]
    \node[roundnode-1] (x1) at (0,0) {$t_1$};
    \node[roundnode-1] (x2) at (0,-1) {$t_2$};
    \node[squarednode-2] (G11) at (1.5,0) {$G$};
    \node[squarednode-2] (G12) at (1.5,-1) {$G$};
    \node[roundnode-2] (f11) at (3,0) {$\varphi^1_1$};
    \node[roundnode-2] (f12) at (3,-1) {$\varphi^1_2$};
    \draw[thick, ->] (x1) -- (G11);
    \draw[thick, ->] (x2) -- (G12);
    \draw[thick, ->] (G11) -- (f11);
    \draw[thick, ->] (G12) -- (f12);

    \node[squarednode] (sum-1) at (4.5, -0.5) {$\Sigma\rightarrow W$};
    \node[roundnode-1] (g11) at (6, 0) {$g^1_1$};
    \node[roundnode-1] (g12) at (6, -1) {$g^1_2$};
    \node[squarednode-2] (G21) at (7.5,-0) {$G$};
    \node[squarednode-2] (G22) at (7.5,-1.) {$G$};
    \draw[thick,->] (f11) -- (sum-1);
    \draw[thick,->] (f12) -- (sum-1);
    \draw[thick,->] (sum-1) -- (g11);
    \draw[thick,->] (sum-1) -- (g12);
    \draw[thick,->] (g11) -- (G21);  \draw[thick,->] (g12) -- (G22); 

    \node[roundnode-2] (f21) at (9,-0) {$\varphi^2_1$};
    \node[roundnode-2] (f22) at (9,-1) {$\varphi^2_2$};
    \draw[thick,->] (G21) -- (f21); \draw[thick,->] (G22) -- (f22);

    \node[squarednode] (sum-2) at (10.5, -0.5) {$\Sigma\rightarrow W$};
    \draw[thick,->] (f21) -- (sum-2);
    \draw[thick,->] (f22) -- (sum-2);
    \node[roundnode-1] (g21) at (12,-0.5) {$g^2_1$};
    \draw[thick,->] (sum-2) -- (g21);
    \node[circle, fill=mypurple!50] (y) at (13.5,-0.5) {$y$};
    \draw[thick,->] (g21) -- (y);
\end{tikzpicture}
\caption{Deep latent force model leveraging RFFs (DLFM-RFF)}   \label{fig:subfig-dlfm-rff}
\end{subfigure}

\begin{subfigure}{\textwidth}
\centering
\vspace{1.1em}
\begin{tikzpicture}[scale=0.85, roundnode-1/.style={circle, fill=green!20, font=\small}, roundnode-2/.style={circle, fill=orange!20, font=\small}, 
squarednode/.style={rectangle, draw=black!50,font=\small}, squarednode-2/.style={rectangle, fill=red!20, draw=black!50,font=\small}
]
    \node[roundnode-1] (x1) at (0,0) {$t_1$};
    \node[roundnode-1] (x2) at (0,-1) {$t_2$};
    \node[squarednode-2] (G11) at (1.5,0) {$G$};
    \node[squarednode-2] (G12) at (1.5,-1) {$G$};
    \node[roundnode-2] (f11) at (3,0) {$f^1_1$};
    \node[roundnode-2] (f12) at (3,-1) {$f^1_2$};
    \draw[thick, ->] (x1) -- (G11);
    \draw[thick, ->] (x2) -- (G12);
    \draw[thick, ->] (G11) -- (f11);
    \draw[thick, ->] (G12) -- (f12);

    \node[squarednode] (sum-1) at (4.5, -0.5) {$\Sigma$};
    \node[roundnode-1] (g11) at (6, 0) {$g^1_1$};
    \node[roundnode-1] (g12) at (6, -1) {$g^1_2$};
    \node[squarednode-2] (G21) at (7.5,-0) {$G$};
    \node[squarednode-2] (G22) at (7.5,-1.) {$G$};
    \draw[thick,->] (f11) -- (sum-1);
    \draw[thick,->] (f12) -- (sum-1);
    \draw[thick,->] (sum-1) -- (g11);
    \draw[thick,->] (sum-1) -- (g12);
    \draw[thick,->] (g11) -- (G21);  \draw[thick,->] (g12) -- (G22); 

    \node[roundnode-2] (f21) at (9,-0) {$f^2_1$};
    \node[roundnode-2] (f22) at (9,-1) {$f^2_2$};
    \draw[thick,->] (G21) -- (f21); \draw[thick,->] (G22) -- (f22);

    \node[squarednode] (sum-2) at (10.5, -0.5) {$\Sigma$};
    \draw[thick,->] (f21) -- (sum-2);
    \draw[thick,->] (f22) -- (sum-2);
    \node[roundnode-1] (g21) at (12,-0.5) {$g^2_1$};
    \draw[thick,->] (sum-2) -- (g21);
    \node[circle, fill=mypurple!50] (y) at (13.5,-0.5) {$y$};
    \draw[thick,->] (g21) -- (y);
\end{tikzpicture}
\caption{Our deep latent force model with VFRF (DLFM-VFRF)}   \label{fig:subfig-dlfm}
\end{subfigure}
\caption{A conceptual illustration of how our model (\ref{fig:subfig-dlfm}) differs from the IDDGP (\ref{fig:subfig-idgp}) and the DLFM-RFF (\ref{fig:subfig-dlfm-rff}). Compared to (\ref{fig:subfig-idgp}), our model additionally applies convolution operators $G$ from the ODEs to each input dimension: $f(t)=\int G(t-\tau)u(\tau)\dif\tau$, where $G(\cdot)$ represents the Green's function and $u(\cdot)$ is a GP prior with \matern\ kernels. Compared to (\ref{fig:subfig-dlfm-rff}) using RFFs $\varphi(\cdot)$ for low-rank covariance matrix approximation and making inference over weights $W$, our model uses Fourier features derived from applying linear transformations to GPs and make inference in an inter-domain way. For a high-level comparison with other models, see Fig.~\ref{fig:model-comparison}.}
\label{fig:model-structure}
\end{figure*}
\paragraph{Model interpretation} The Green's function $G(\cdot)$, determined by the system's dynamics, serves as a signal filter. It effectively acts as a low-pass filter described by the ODE \eqref{eq:ode-ab}, with $\gamma$ representing the ``cutoff frequency''. Mathematically, the convolution operator $G(\cdot)$ of the modified ODE will behave like the Dirac delta function in \eqref{eq:f(t)} as $\alpha=1,\beta\rightarrow0^{+}$ (i.e.,$\gamma\rightarrow +\infty$), causing $f(t)$ to closely replicate $u(t)$. Fig.~\ref{fig:basis-functions}, left, illustrates this behaviour. We use the \matern-$\tfrac{1}{2}$ kernel for the covariance $k(t,t')$ of the latent force. The figure shows this covariance function (dashed) and two LFMs covariance functions (solid) $k_{f}(t,t')$ with different $\gamma$ values. The LFM kernel reverts to the latent force kernel as $\gamma$ increases. However, the conventional LFM kernel without the ODE modification, i.e., \eqref{eq:lfm-equation} \citep{fastkernel} will get flattened since the corresponding Green's function $\exp{(-\gamma t)}$ does not effectively mimic a valid Dirac delta function.

\subsection{Variational Fourier Response Features}
\label{sec:VFRFs}

Building upon the modified ODE described by \eqref{eq:ode-ab}, we introduce a spectral approximation for the LFMs within the inter-domain GP framework. The latent force $u$ is initially projected onto the Fourier basis entries $\phi_m$ as defined in \eqref{eq:fourier-basis},  yielding its spectral representations
\begin{equation}
    v_m = \langle \phi_m,  u\rangle_{\mtc{H}},\quad m=0,1,\ldots, 2M.
    \label{eq:v_m}
\end{equation}
The projected inducing variables $v_m$ are collected as $\mbf{v}=[v_m]_{m=0}^{2M}\in\R{2M+1}$. 

By the closure of  GPs under linear operations, the output $f$ and the projection $\mbf{v}$ share a joint augmented GP prior. The covariance matrix of inducing variables $\mathrm{Cov}[\mbf{v}, \mbf{v}]$ has  a low-rank-plus-diagonal structure if inducing frequencies $\mathbf{z}=[z_m]_{m=1}^M$ are harmonic on $[a,b]$, i.e., $z_m=\frac{2\pi m}{b-a}$, facilitating faster posterior computation \citep{vffs}. For a given input $t$, the cross-covariance of the output process $f$ and the inducing variable $v_m$ is computed as
\begin{equation}
    \mathrm{Cov}[f(t), v_m]  = 
    \int_{-\infty}^t G(t-\tau) \cdot
    \langle k(\tau,\cdot), \phi_m(\cdot)\rangle_{\mathcal{H}}\dif\tau,
    \label{eq:cov-fv}
\end{equation}
where we take advantage of the linearity and calculate the expectation over $u(\cdot)$. The reproducing property of \matern\ RKHS ensures that the inner product $\langle k(\tau,\cdot), \phi_m(\cdot)\rangle_{\mathcal{H}}$ results in well-defined sinusoidal functions within the interval $\tau\in[a,b]$. Therefore, we can derive the RKHS Fourier features for LFMs on $t\in[a,b]$ as follows:
\begin{equation}
\begin{aligned}
    \mathrm{Co}&\mathrm{v}[f(t), v_m] = \int_{-\infty}^t G(t-\tau)\phi_m(\tau)\dif\tau \\
    & =
    \begin{cases}
        \frac{\cos(z_i(t-a)+\theta)}{\beta\sqrt{z_i^2+\gamma^2}} + \xi_i 
        & i =0,\ldots,M,\\
        \addls
         \frac{\sin(z_i(t-a)+\theta)}{\beta\sqrt{z_i^2+\gamma^2}} + \xi_i
        & i =M+1,\ldots, 2M,
    \end{cases}
\end{aligned}
\label{eq:gkm-feature}
\end{equation}
where $z_0=0$, cutoff frequency $\gamma=\frac{\alpha}{\beta}$, phase shift $\theta=-\arctan(\frac{z_i}{\gamma})$, and $\xi_i$ represents an exponential decay term. Since the integration variable $\tau$ ranges from negative infinity and the inner product $\langle k(\tau,\cdot), \phi(\cdot, z)\rangle_{\mathcal{H}}$ has different expressions beyond $\tau\in[a,b]$,  the covariance of $f$ and $v_m$ for $t\in\RR$ emerges as a continuous piece-wise function (see Fig.~\ref{fig:basis-functions} right). The detailed derivation and complete expressions of $\text{Cov}[f(t),v(z)]$ for \matern-$\tfrac{1}{2}/\frac{3}{2}/\frac{5}{2}$ are given in Appendix~\ref{app:mkg} and illustrated in Fig.~\ref{fig:basis-function32} and \ref{fig:basis-function52}.

The derived inter-domain features \eqref{eq:gkm-feature} reflect the filtering effect of the system, i.e., how the ODE adaptively transforms the frequency components of latent forces to the output through amplitude attenuation and phase shift. By analogy with the frequency response of linear systems, we name the obtained Fourier features from RKHS as \emph{Variational Fourier Response Features} (VFRFs). Fig.~\ref{fig:basis-functions}, right, depicts the VFRFs of the LFM from the left subplot ($\gamma=4$, solid red) and the VFFs of the corresponding latent force (dash blue). They show that the VFRFs are learnable basis functions that adjust both the amplitude and the phase according to the input frequencies and the ODE parameters. Particularly, the system will allow nearly all frequency components of the input process to pass through as $\gamma\rightarrow +\infty$. Under this condition, the VFRFs converge to the VFFs. We would like to emphasise here that the features derived from \eqref{eq:cov-fv} can apply to more general inter-domain GPs with other linear transformations $G(\cdot)$, not just limited to LFMs. Moreover, for stable dynamical systems governed by higher-order ODEs within the LFM framework, the derivation of the VFRFs described above can be readily extended by using corresponding Green's functions. 

\subsection{Deep LFMs with VFRFs}
\label{sec:DLFM}

DLFMs extend the concept of shallow LFMs by stacking them in a non-parametric cascade, similar to DGPs. This hierarchical structure allows DLFMs to model the non-stationarities present in nonlinear dynamical systems. In this section, we detail the construction of a hierarchical composition of $L$ LFMs within the framework of variational DGPs, each governed by the modified ODE and enhanced with VFRFs for variational approximation of the posterior. Fig.~\ref{fig:model-structure} gives a conceptual illustration of how our proposed DLFM differs from a DGP. We leverage the layer-wise Monte Carlo technique in doubly stochastic variational inference \citep{dgp2017} to allow functional samples to propagate through the compositional architecture efficiently. 

The first layer of a DLFM processes a $D^{0}$-dimensional input $\mbf{t}=[t_d]_{d=1}^{D^{0}}$ and outputs a $D^{1}$-dimensional independent process $\bds{g}^1(\mbf{t})=[g^1_r(\mbf{t})]_{r=1}^{D^1}$ (the superscripts indicate the layer index). To extend the application of VFRFs to multidimensional inputs, we follow \cite{vffs} to employ additive LFM kernels for each output dimension, i.e., each output dimension $g^1_r(\mbf{t})$ is modelled as $g^1_r(\mbf{t}) = \sum_{d=1}^{D^{0}}f^1_d(t_d)$, where $\{f^1_d\}_{d=1}^{D^{0}}$ are LFMs with ODE-induced covariance functions. In this work, we assume the LFMs $f^1_d$ are independent, but this assumption can be relaxed by allowing them to share the same latent forces, which can lead to more complex kernels for the outputs. 

Following the construction of a single-layer LFM, each $g^1_r(\mbf{t})$ is equipped with a set of $M$ inducing frequencies $\mbf{Z}^{0}\in\R{M\times D^{0}}$ and corresponding inducing variables $\mbf{V}^{1}=[v^1_{m,d}]^{2M+1, D^{0}}_{m=1,d=1}$. These variables are created by the RKHS projection $v^1_{m,d}=\langle u^1_d, \phi_m\rangle_{\mtc{H}}$. Therefore, the covariance functions necessary for sparse variational inference are given by
\begin{equation*}
\begin{aligned}
    \mathrm{Cov}[g^1_r(\mbf{t}), v^1_{m,d}] &= \int_{-\infty}^{t_d} G^1_d(t_d-\tau)\phi_m(\tau)\dif\tau, \\
    \mathrm{Cov}[v^1_{m,d},v^1_{m',d'}] &= 0\ \ (d\neq d').
\end{aligned}
\end{equation*}
Assuming a variational distribution $q(\mbf{V}^1)$, the approximate posterior $q(\bds{g}^1\mid\mbf{t})$ of the first layer is derived by substituting $k(\mbf{x},\mbf{Z})$ and $\mbf{K_{ZZ}}$ in \eqref{eq:svgp-post} with the expressions of VFRFs. Samples from this approximate posterior are drawn using the re-parameterisation trick \citep{kingma2015variational}. 

Given a training dataset with inputs $[\mbf{t}_i]_{i=1}^N$ and targets $[\mbf{y}_i]_{i=1}^N$, the training of DLFMs involves maximising the average ELBO over a mini-batch $\mtc{B}$:
\begin{equation*}
\begin{aligned}
    \mathrm{ELBO} =&  
    \frac{1}{|\mtc{B}|}\sum_{i\in\mtc{B}}\mathbb{E}_{q(\mbf{g}^L_i\mid\mbf{t}_i)}\log\left[p(\mbf{y}_i\mid \mbf{g}^L_i)\right] \\
    & - \frac{1}{N}\sum_{l=1}^L\mathrm{KL}\left[q(\mbf{V}^l)\mid\mid p(\mbf{V}^l)\right],
\end{aligned}
\end{equation*}
where $\mbf{V}^l$ are the inducing variables of the $l$-th layer, and the collection $\{\mbf{g}^l\}^L_{l=1}$ denotes the output random variables at hidden layers. The output of each layer serves as the input for the subsequent layer, creating a chain of dependencies where the posterior of each layer is computed based on the propagated samples $\hat{\mbf{g}}_i^l\sim q(\mbf{g}_i^l\mid\hat{\mbf{g}}_i^{l-1})$. The predictive distribution $q(\mbf{y}_*)=\int p(\mbf{y}_*\mid\mbf{g}^L_*)q(\mbf{g}^L_*)\dif\mbf{g}^L_*$ at test location $\mbf{t}_*$ follows a similar layer-wise procedure, where $q(\mbf{g}^L_*)$ is a Gaussian mixture of $S$ hidden-layer samples: $q(\mbf{g}_*^L)\approx\frac{1}{S}\sum_{s=1}^S q(\mbf{g}_*^L\mid\hat{\mbf{g}}_*^{(s)^{L-1}})$.

\section{RELATED WORK}

LFMs present a physically-inspired approach to combining data-driven modelling with differential equations \citep{lfms}. \cite{alvarez2010efficient} further proposed variational inducing functions to handle non-smooth latent processes within convolved GPs \citep{alvarez2011computationally}. Our model builds upon LFMs and DGPs \citep{dgp2017}. Recently, various approximate inference methods have been explored for DGP-based models,  which are generally categorised into variational inference techniques \citep{dgp2017, dgp2019importance, dgp2020beyond} and Monte Carlo approaches\citep{dgp2018}. 

As outlined in Section~\ref{sec:Intro},  RFFs \citep{rahimi2007random} and VFFs \citep{vffs} have recently been incorporated into GP models. RFFs were used in shallow LFMs models to approximate covariance matrices \citep{fastkernel} and expanded to a deeper architecture \citep{dlfm2021,dlfm2023}. VFFs were once integrated with harmonizable mixture kernels in shallow GP models \citep{pmlr-v89-shen19c}. While related to these studies, our approach primarily uses features similar to VFFs within the scope of inter-domain GPs \citep{inter-domain-gp2009,van2020framework}. Unlike RFF-based DGP models, which often modify the original covariance functions by introducing a fully parametric variational distribution over random frequencies, our model preserves the integrity of the original kernel forms and approximates the DGP posterior directly. Another closely related work is Inter-Domain DGPs (IDDGPs) \citep{iddgp2020}, which employ fixed VFFs without ODEs. We provide an illustrative plot of IDDGP in Fig.~\ref{fig:subfig-idgp}. In contrast, our model extends the compositional inter-domain GPs by integrating ODEs to provide trainable, physics-informed RKHS Fourier features. Fig.~\ref{fig:model-comparison} in Appendix~\ref{app:model-comparison} illustrates a high-level comparison of our work with related studies.

\section{EXPERIMENTS}
\label{sec:EXP}

This section presents experiments designed to evaluate our model using VFRFs. We begin by examining the approximation quality of shallow LFMs with VFRFs and RFFs on synthetic data. We then evaluate our model on a highly non-stationary speech signal dataset and benchmark regression tasks, comparing it to various baselines in both cases.\footnote{Our code is publicly available in the repository: \texttt{https://github.com/shixinxing/LFM-VFF}} 
\begin{figure}[t]
    \centering
    \includegraphics[width=\linewidth]{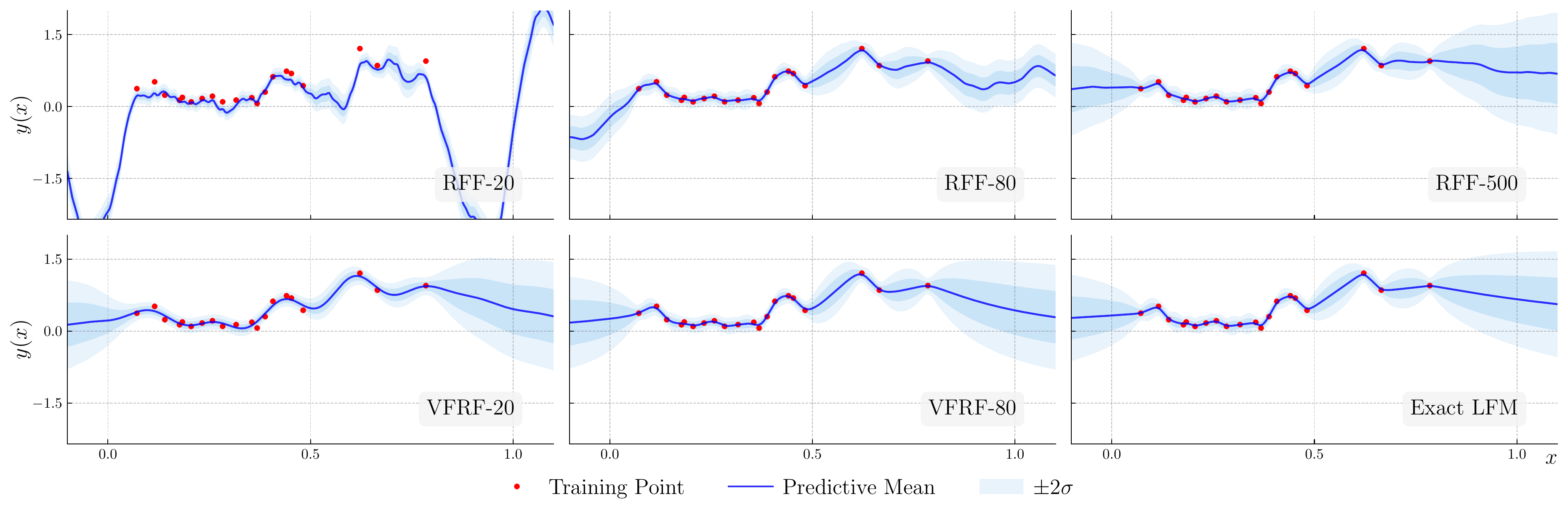}
    \caption{Illustrative example of \matern-$\tfrac{1}{2}$ LFM posteriors with VFRFs / RFFs. The model's feature is indicated at the lower right. \textbf{Top row}: predictive posteriors of 20, 80, and 500 RFFs. \textbf{Bottom row}: predictive posteriors of 20 and 80 inducing frequencies and an exact LFM. Noisy observations are marked with red dots,  posterior predictive means with blue lines, and uncertainty (one or two standard deviations) with varying shades of blue. In this example, VFRFs show a better approximation to the true posterior, whereas RFFs indicate variance underestimation with fewer features.}
    \label{fig:vff_vs_rff}
\end{figure}
\begin{figure}[t]
    \centering
    \includegraphics[width=\linewidth]{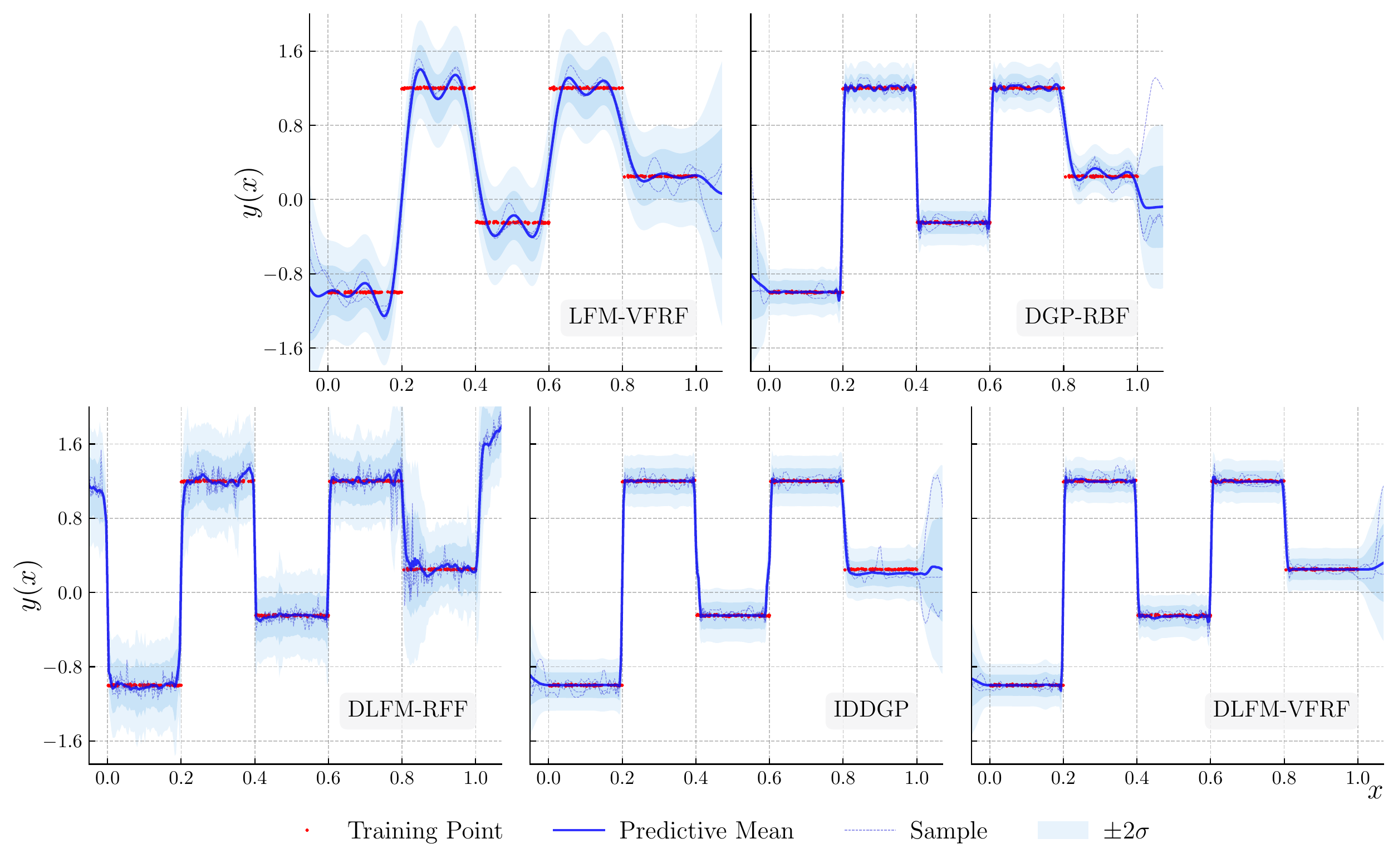}
    \caption{Posterior predictive distribution comparison of different models on data points from a noisy multi-step function. The models and the features used are noted at the bottom right of each subplot. The dashed lines are samples from the predictive distributions. The experiment uses two layers for deep models and \matern-$\tfrac{3}{2}$ kernels except for the DGP (upper left) and DLFM-RFF (lower left) that use RBF kernels. All models are trained with 20 inducing points/Fourier features per layer. The DLFM models with VFRFs perform best among the models.}
    \label{fig:synthetic}
\end{figure}
\subsection{Synthetic Datasets}

We first evaluate the shallow LFMs and DLFMs using the proposed VFRFs on two synthetic datasets, respectively.

\subsubsection{Posterior Approximation for Shallow LFMs}

VFRFs and RFFs both leverage Fourier features to facilitate approximate inference in LFMs. In Fig.~\ref{fig:vff_vs_rff}, we compare the approximation quality of VFRFs and RFFs in a regression task using models with a \matern-$\frac{1}{2}$ kernel. The kernel's parameters and the noise variance are initially optimised by maximising the marginal likelihood of an exact LFM and then fixed across all models. The frequencies of RFFs are sampled from the corresponding Cauchy distribution of the kernel (detailed in Appendix~\ref{app:fastkernel}). 

Fig.~\ref{fig:vff_vs_rff} shows that the model using 20 VFRFs has already fitted the data points reasonably well. Increasing VFRFs to 80 fills in the details of the region with more observations, and the approximate predictive posterior is quite close to the exact one. The same number of RFFs yields a poorly fitted predictive mean and tends to underestimate the variance in different regions of the input space, which is a phenomenon known as \emph{variance starvation} \citep{var-starvation2018}. From the top row of Fig.~\ref{fig:vff_vs_rff}, we can observe that achieving a comparable approximation requires more RFFs than VFRFs due to the heavy-tailed spectral density of the \matern-$\frac{1}{2}$ kernels used in RFFs.

\begin{figure*}[htb]
  \centering
  \begin{subfigure}[b]{0.65\textwidth}
    \includegraphics[width=\linewidth]{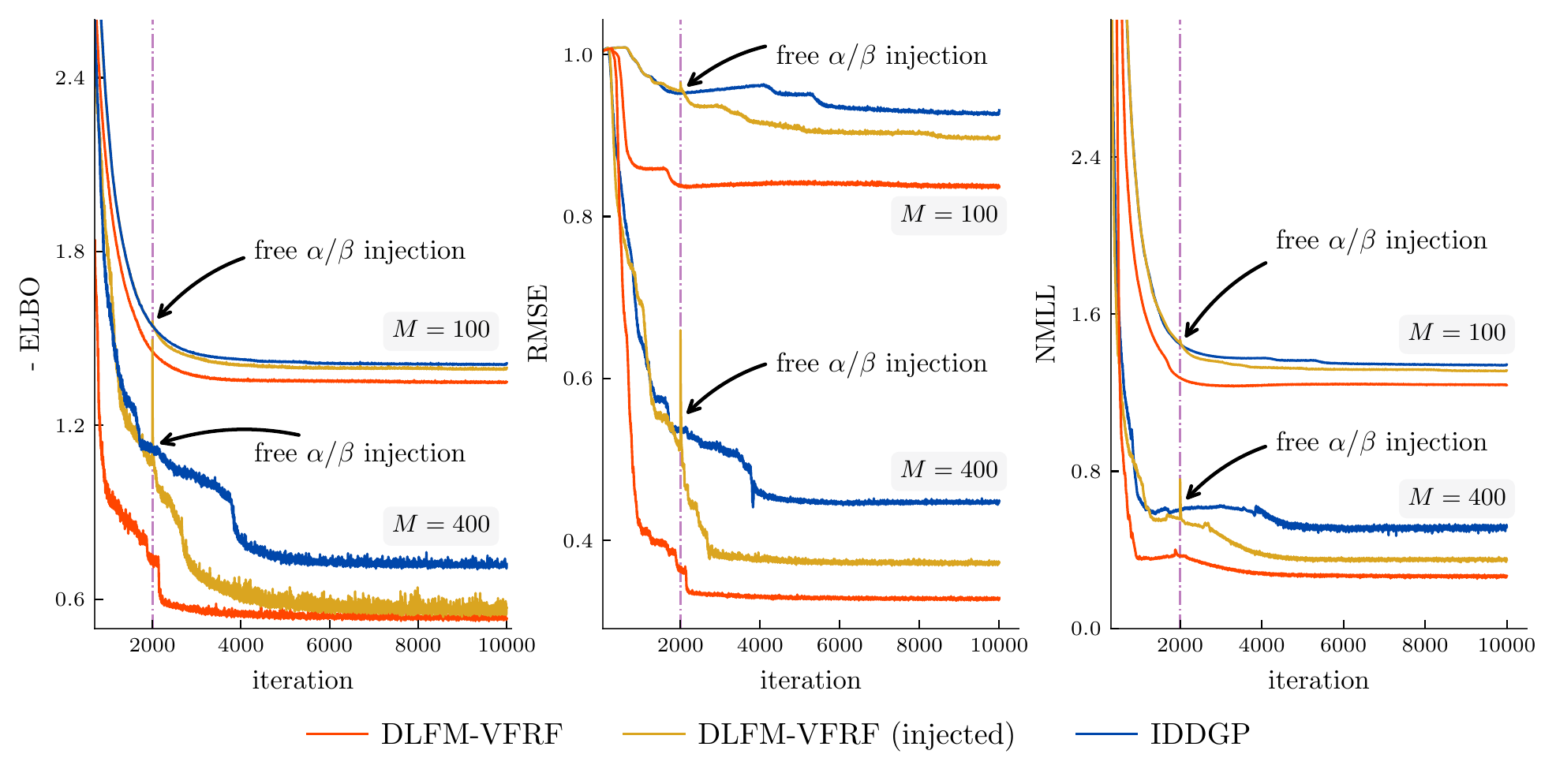}
    \caption{Learning progression of DLFM-VFRF and IDDGP}
    \label{subfig:dlfm-vs-iddgp}
  \end{subfigure}
  \begin{subfigure}[b]{0.34\textwidth}
    \includegraphics[width=\linewidth]{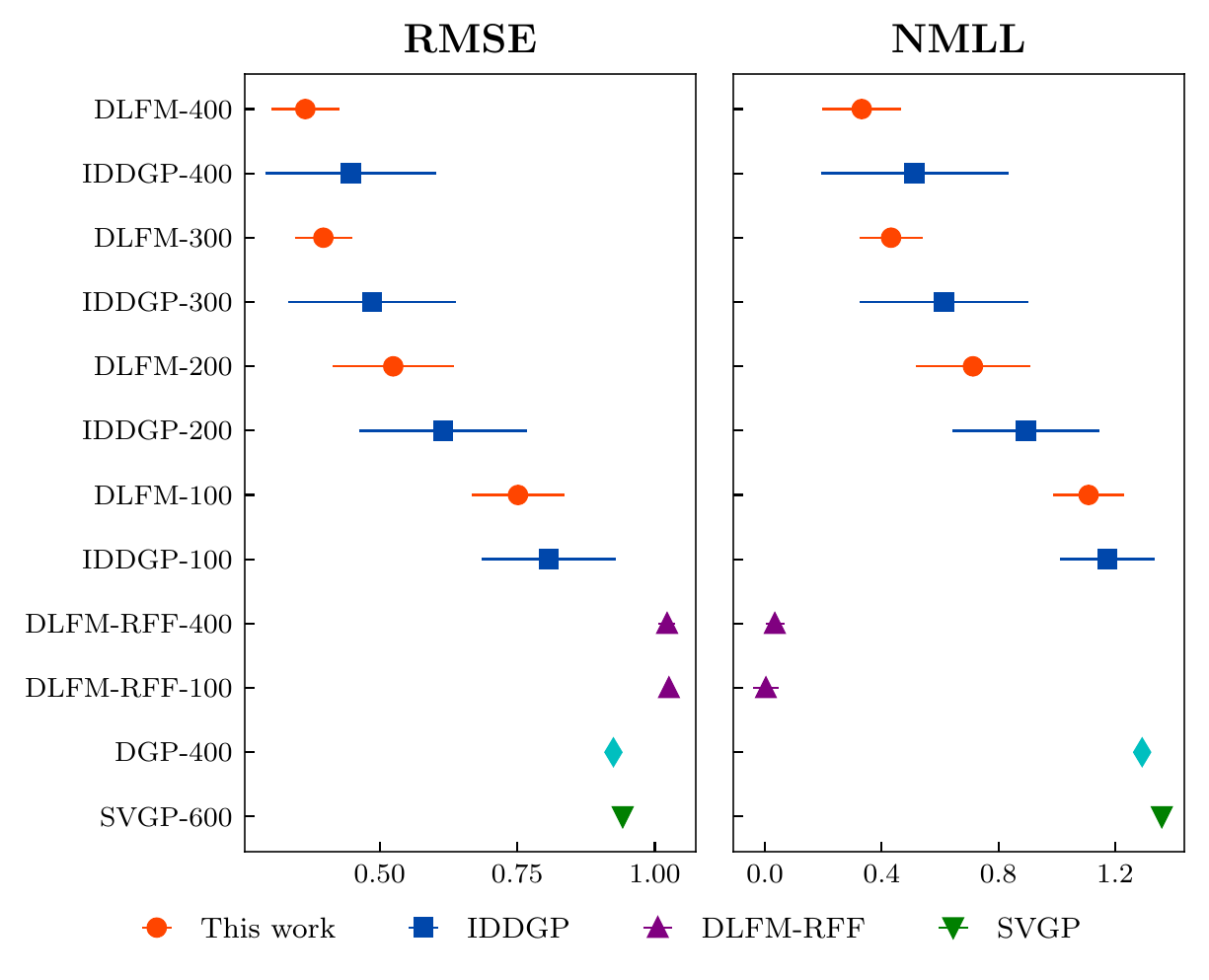}
    \vspace{0.5pt}
    \caption{Standardised RMSE and NMLL}
    \label{subfig:timit_results}
  \end{subfigure}
  \caption{\textbf{(a)} Learning progression of DLFMs and IDDGPs with $M$ inducing frequencies on the TIMIT dataset, presented in negative ELBO, test average RMSE and NMLL. The DLFM in yellow maintains fixed $\beta=10^{-6}$ throughout the first 2000 training iterations, after which $\alpha/\beta$ are allowed to vary. The DLFMs in red employ trainable ODE parameters from the start. The DLFM-VFRFs consistently outperform the IDDGPs; 
  \textbf{(b)} Mean standardised RMSE and NMLL with the standard deviations (over 10 random seeds) for models employing varying numbers of inducing frequencies. The numbers following the hyphen in the y-axis labels indicate the number of inducing frequencies/points. A lower value (to the left) indicates better performance.}
  \label{fig:timit}
\end{figure*}

\subsubsection{Multi-step Function for Deep Structures} 

To further evaluate our DLFM-VFRF's performance against other models, we conduct tests on a synthetic multi-step function (as shown in Fig.~\ref{fig:synthetic}). This task is challenging for shallow GP models due to the need to capture global structures in highly non-stationary data \citep{iddgp2020}. 

Although equipped with VFRFs, our single-layer LFM (upper left) struggles to fit the non-stationarity with a stationary kernel. In contrast, the models with compositional layers exhibit better performance. The DLFM-RFF (lower left) \citep{dlfm2021} generates high-frequency, wiggly posterior predictive samples, resulting in an easily over-fitting model struggling to seize the slow-changing trend in the data. DGP-RBF (upper left), IDDGP (lower middle), and our DLFM-VFRF (lower right) offer smoother samples from the posterior distributions. Due to the VFRFs' flexibility to capture the global data structure, our DLFM-VFRF outperforms both the DGP-RBF with local inducing points and the IDDGP, especially inside $[0.8, 1]$. Our model provides a more accurate predictive mean throughout the steps and at abrupt step transitions and demonstrates narrower confidence intervals, indicating a better uncertainty calibration. In the plot, GP models based on function-space inference tend to revert to prior distributions outside the observed data range, displaying wide uncertainty bands. In contrast, the DLFM-RFF yields relatively more confident non-zero predictions in these areas. 

To measure the performance quantitatively, we conducted additional experiments to train five independent copies of the IDDGP and our model. We summarise the Root Mean Square Error (RMSE) and Negative Marginal Log-Likelihood (NMLL) on the test points in the following Table~\ref{tab:iddgp-vs-dlfm-vfrf}. Additionally, the outputs of the intermediate layers for DGP, IDDGP, and our model are shown in Fig.~\ref{fig:hidden-layers} in Appendix~\ref{app:experiment-details}.
\begin{table}[h]
    \centering
    \caption{Performance of IDDGP and our DLFM-VFRF on fitting the multi-step function over five runs (lower is better).}
    \begin{tabular}{ccc}
    \toprule
    Model & RMSE & NMLL \\
    \midrule
    IDDGP         & 0.107 $\pm$ 0.015 & -1.051 $\pm$ 0.131 \\
    Ours     & \textbf{0.095 $\pm$ 0.010}  & \textbf{-1.304 $\pm$ 0.120}\\
    \bottomrule
    \end{tabular}
    \label{tab:iddgp-vs-dlfm-vfrf}
\end{table}

\subsection{TIMIT Speech Signals} 

We apply our model to a regression task on the TIMIT dataset, a speech recognition resource previously used by \cite{iddgp2020}, to explore the capability of GP-based models in handling complex, non-stationary data. The dataset features rapidly changing audio waves, posing significant challenges for shallow GP models reliant on local approximation. Initially, we apply a moving average filter to smooth the audio waves and select the first 10,000 data points for our analysis, reserving 30\% as test data. Our method uses the \matern-$\frac{3}{2}$ kernels.

One of our goals is to evaluate how the performance of IDDGPs and DLFMs varies with the number of global Fourier features and the effect of the ODE parameters $\alpha$ and $\beta$ on the learning process. Fig.~\ref{subfig:dlfm-vs-iddgp} illustrates the progressions of performance metrics, e.g., test RMSE and NMLL using 100 and 400 inducing frequencies. The yellow lines of DLFMs align closely with the IDDGP's blue line during the first 2000 iterations, where DLFMs maintain a fixed small $\beta$ value ($\beta=10^{-6}, \alpha=1$). We use this setting to illustrate how our DLFM-VFRF can replicate the original IDDGPs as expected when $\beta\rightarrow 0^+$. After $2000$ iterations, we allow optimisation of $\alpha$ and $\beta$, leading to subsequent improvements in the testing metrics, and suggesting continuous learning with ODE-based Fourier features. Additionally, optimising all parameters from the beginning (red lines) yields the best results across various setups. Fig.~\ref{subfig:timit_results} compares the performance of different models with all parameters optimised from the beginning. It is unsurprising that increasing the number of inducing frequencies typically results in better performance. The results reveal that while RFF-based DLFMs exhibit the lowest NMLL, they show the highest RMSE, reflecting a lack of precision on test data points. DLFMs equipped with VFRFs consistently surpass both the DGP with local inducing points and the IDDGPs in terms of both RMSE and NMLL, highlighting our model's enhanced ability to accurately capture the global structure and non-stationarity of the data.

\paragraph{Running Time Comparison}

Theoretically, the extra running time of our model compared to IDDGPs mainly lies in the more complex forward computation on covariance entries and the backward gradient update on the extra ODE parameters. We record the wall-clock running time (per iteration) of models with the number of inducing frequencies ranging from 100 to 400 in Table~\ref{tab:running-time} below. The results are averaged over five runs, and we exclude the standard deviations as they are quite small.
\begin{table}[ht]
    \centering
    \caption{Wall-clock training time of IDDGPs and our DLFM-VFRFs with different numbers of inducing frequencies $M$.}
    \begin{tabular}{lcccc}
    \toprule 
    Model/$M$ & 100 & 200 & 300 & 400 \\
    \midrule
    IDDGP-$M$ & 0.279s & 0.633s & 1.138s & 1.891s \\
    Ours-$M$  & 0.313s & 0.686s & 1.207s & 1.977s \\
    \bottomrule
    \end{tabular}
    \label{tab:running-time}
\end{table}

From Table~\ref{tab:running-time}, we observe that our model incurs slightly higher runtime overhead compared to IDDGP. Despite this, the runtime difference compared with IDDGP with VFFs remains acceptable even with 400 inducing features, especially given the improved flexibility and modelling capacity of our approach. The runtime gap might be reduced with some computational optimization techniques (e.g., JIT) implemented.

\subsection{UCI Regression Benchmarks}
\begin{figure}[th]
    \centering
    \includegraphics[width=\linewidth]{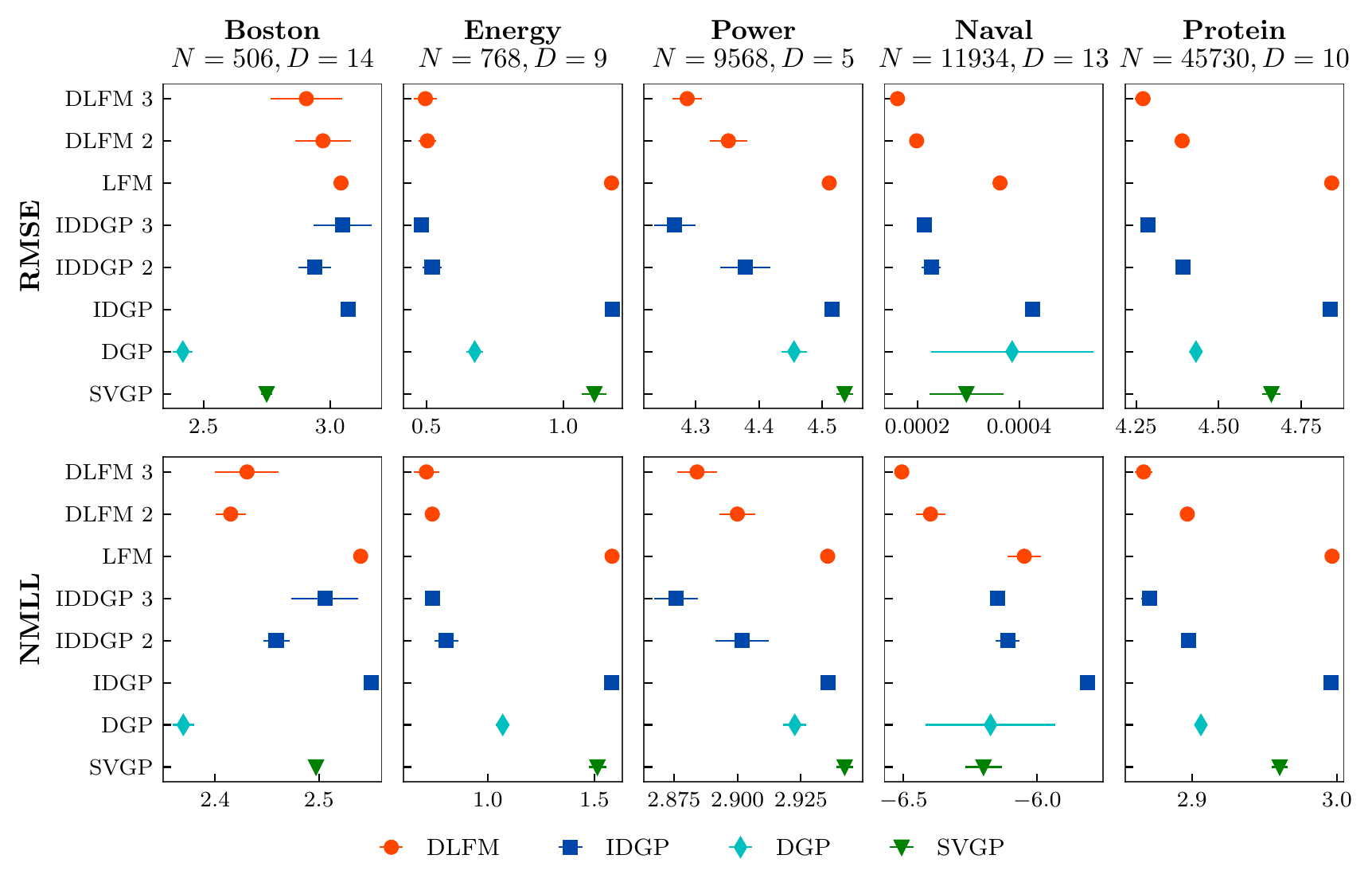}
    \caption{Regression test RMSE and NMLL results on UCI datasets, averaged over 10 random seeds. Lower values (to the left) indicate better performance. Model names include the number of layers.}
    \label{fig:uci-results}
\end{figure}
To demonstrate the versatility and effectiveness of our model on domain-agnostic real-world data, we conduct evaluations on five diverse regression datasets from the UCI Machine Learning Repository \citep{UCI}. These datasets vary in size and feature dimensionality, allowing us to test the model's adaptability across different scenarios (see Fig.~\ref{fig:uci-results}). Consistent with standard practice \citep{dgp2017}, our regression tasks involve multivariate inputs and a univariate target. We reserve 10\% of each dataset for testing, normalise the inputs to the range $[0,3]$, and standardise the outputs based on the mean and standard deviation of the training set (these transformations are then reversed for evaluation). Following the setups in \cite{iddgp2020} and \cite{dlfm2021}, all models run with \matern-$\frac{3}{2}$ kernels and employ 20 inducing points or frequencies. We employ three output dimensions per layer. We maintained the same experimental settings and initialisation across all tests. The figure illustrates that our models achieve comparable performance to the baselines. Notably, our models with two layers outperform IDDGP counterparts on the Energy, Power, and Naval datasets. We also observe that increasing the number of layers generally enhances the model’s representational capacity, resulting in improved performance.

\section{CONCLUSION}

In this work, we adapt VFFs to the latent force framework, which inherently involves convolution operators with Green's functions. This adaptation introduces flexibility in modelling dynamics while preserving computational traceability. By introducing trainable parameters in the Green's function, we provide a mechanism for dynamically adjusting the inter-domain features. We further employ the inter-domain Fourier features in hierarchical LFMs. Our empirical evaluations across various datasets demonstrate that our model extends inter-domain GPs with RKHS Fourier features and has enhanced their modelling capacity for non-stationary and global structures.

\paragraph{Limitations and Future Work} 

The current experiments are only based on models from first-order ODEs. Besides, computing the piece-wise VFRFs at intermediate layers may result in extra computational costs. Future work will focus on developing a normalization method at intermediate layers to accelerate inference and on extending our model's use to other challenging machine-learning tasks requiring the integration of specific domain knowledge of higher-order ODEs. Extending DLFMs to incorporate other recently proposed Fourier features, such as those in \cite{cheema2024integrated}, represents a promising direction.

\subsubsection*{Acknowledgements}

We are grateful for the computational resources from UoM CSF3, UK, and appreciate the insightful discussions with Xiaoyu Jiang. X.S. was supported by the UoM-CSC Joint Scholarship.

\bibliography{refs}

\begin{thebibliography}{}

\bibitem[Alvarez et~al., 2009]{lfms}
Alvarez, M., Luengo, D., and Lawrence, N.~D. (2009).
\newblock Latent force models.
\newblock In {\em Artificial Intelligence and Statistics}, pages 9--16. PMLR.

\bibitem[{\'A}lvarez et~al., 2010]{alvarez2010efficient}
{\'A}lvarez, M., Luengo, D., Titsias, M., and Lawrence, N.~D. (2010).
\newblock Efficient multioutput gaussian processes through variational inducing kernels.
\newblock In {\em Proceedings of the Thirteenth International Conference on Artificial Intelligence and Statistics}, pages 25--32. JMLR Workshop and Conference Proceedings.

\bibitem[Alvarez and Lawrence, 2011]{alvarez2011computationally}
Alvarez, M.~A. and Lawrence, N.~D. (2011).
\newblock Computationally efficient convolved multiple output gaussian processes.
\newblock {\em The Journal of Machine Learning Research}, 12:1459--1500.

\bibitem[Cheema and Rasmussen, 2024]{cheema2024integrated}
Cheema, T.~M. and Rasmussen, C.~E. (2024).
\newblock Integrated variational fourier features for fast spatial modelling with gaussian processes.
\newblock {\em Transactions on Machine Learning Research}.

\bibitem[Cutajar et~al., 2017]{cutajar2017random}
Cutajar, K., Bonilla, E.~V., Michiardi, P., and Filippone, M. (2017).
\newblock Random feature expansions for deep gaussian processes.
\newblock In {\em International Conference on Machine Learning}, pages 884--893. PMLR.

\bibitem[Damianou and Lawrence, 2013]{dgp2013}
Damianou, A. and Lawrence, N.~D. (2013).
\newblock Deep gaussian processes.
\newblock In {\em Artificial Intelligence and Statistics}, pages 207--215. PMLR.

\bibitem[Dua and Graff, 2019]{UCI}
Dua, D. and Graff, C. (2019).
\newblock {UCI} machine learning repository.
\newblock Irvine, CA: University of California, School of Information and Computer Sciences.

\bibitem[Durrande et~al., 2016]{durrande2016detecting}
Durrande, N., Hensman, J., Rattray, M., and Lawrence, N.~D. (2016).
\newblock Detecting periodicities with gaussian processes.
\newblock {\em PeerJ Computer Science}, 2:e50.

\bibitem[Duvenaud et~al., 2014]{duvenaud14avoid}
Duvenaud, D., Rippel, O., Adams, R., and Ghahramani, Z. (2014).
\newblock {Avoiding pathologies in very deep networks}.
\newblock In Kaski, S. and Corander, J., editors, {\em Proceedings of the Seventeenth International Conference on Artificial Intelligence and Statistics}, volume~33 of {\em Proceedings of Machine Learning Research}, pages 202--210, Reykjavik, Iceland. PMLR.

\bibitem[Gal and Turner, 2015]{gal2015improving}
Gal, Y. and Turner, R. (2015).
\newblock Improving the gaussian process sparse spectrum approximation by representing uncertainty in frequency inputs.
\newblock In {\em International Conference on Machine Learning}, pages 655--664. PMLR.

\bibitem[Gardner et~al., 2018]{gardner2018gpytorch}
Gardner, J.~R., Pleiss, G., Bindel, D., Weinberger, K.~Q., and Wilson, A.~G. (2018).
\newblock Gpytorch: Blackbox matrix-matrix gaussian process inference with gpu acceleration.
\newblock In {\em Advances in Neural Information Processing Systems}.

\bibitem[Guarnizo and {\'A}lvarez, 2018]{fastkernel}
Guarnizo, C. and {\'A}lvarez, M.~A. (2018).
\newblock Fast kernel approximations for latent force models and convolved multiple-output gaussian processes.
\newblock In {\em 34th Conference on Uncertainty in Artificial Intelligence 2018, UAI 2018}, pages 835--844. Association For Uncertainty in Artificial Intelligence (AUAI).

\bibitem[Havasi et~al., 2018]{dgp2018}
Havasi, M., Hern{\'a}ndez-Lobato, J.~M., and Murillo-Fuentes, J.~J. (2018).
\newblock Inference in deep gaussian processes using stochastic gradient hamiltonian monte carlo.
\newblock {\em Advances in Neural Information Processing Systems}, 31.

\bibitem[Hensman et~al., 2018]{vffs}
Hensman, J., Durrande, N., and Solin, A. (2018).
\newblock Variational fourier features for gaussian processes.
\newblock {\em Journal of Machine Learning Research}, 18(151):1--52.

\bibitem[Hensman et~al., 2013]{gpbigdata2013}
Hensman, J., Fusi, N., and Lawrence, N.~D. (2013).
\newblock Gaussian processes for big data.
\newblock In {\em Proceedings of the Twenty-Ninth Conference on Uncertainty in Artificial Intelligence}, UAI'13, page 282–290, Arlington, Virginia, USA. AUAI Press.

\bibitem[Hensman et~al., 2015]{hensman2015scalable}
Hensman, J., Matthews, A., and Ghahramani, Z. (2015).
\newblock Scalable variational gaussian process classification.
\newblock In {\em Artificial Intelligence and Statistics}, pages 351--360. PMLR.

\bibitem[Kingma et~al., 2015]{kingma2015variational}
Kingma, D.~P., Salimans, T., and Welling, M. (2015).
\newblock Variational dropout and the local reparameterization trick.
\newblock {\em Advances in Neural Information Processing Systems}, 28.

\bibitem[Lawrence et~al., 2006]{lawrence2006modelling}
Lawrence, N., Sanguinetti, G., and Rattray, M. (2006).
\newblock Modelling transcriptional regulation using gaussian processes.
\newblock {\em Advances in Neural Information Processing Systems}, 19.

\bibitem[L{\'a}zaro-Gredilla and Figueiras-Vidal, 2009a]{inter-domain-gp2009}
L{\'a}zaro-Gredilla, M. and Figueiras-Vidal, A. (2009a).
\newblock Inter-domain gaussian processes for sparse inference using inducing features.
\newblock {\em Advances in Neural Information Processing Systems}, 22.

\bibitem[L{\'a}zaro-Gredilla and Figueiras-Vidal, 2009b]{lazaro2009inter}
L{\'a}zaro-Gredilla, M. and Figueiras-Vidal, A. (2009b).
\newblock Inter-domain gaussian processes for sparse inference using inducing features.
\newblock {\em Advances in Neural Information Processing Systems}, 22.

\bibitem[L{\'a}zaro-Gredilla et~al., 2010]{lazaro2010sparsespectrum}
L{\'a}zaro-Gredilla, M., Quinonero-Candela, J., Rasmussen, C.~E., and Figueiras-Vidal, A.~R. (2010).
\newblock Sparse spectrum gaussian process regression.
\newblock {\em The Journal of Machine Learning Research}, 11:1865--1881.

\bibitem[Lindinger et~al., 2020]{dgp2020beyond}
Lindinger, J., Reeb, D., Lippert, C., and Rakitsch, B. (2020).
\newblock Beyond the mean-field: Structured deep gaussian processes improve the predictive uncertainties.
\newblock {\em Advances in Neural Information Processing Systems}, 33:8498--8509.

\bibitem[McDonald and {\'A}lvarez, 2021]{dlfm2021}
McDonald, T. and {\'A}lvarez, M. (2021).
\newblock Compositional modeling of nonlinear dynamical systems with ode-based random features.
\newblock {\em Advances in Neural Information Processing Systems}, 34:13809--13819.

\bibitem[McDonald and {\'A}lvarez, 2023]{dlfm2023}
McDonald, T. and {\'A}lvarez, M. (2023).
\newblock Deep latent force models: Ode-based process convolutions for bayesian deep learning.
\newblock {\em arXiv preprint arXiv:2311.14828}.

\bibitem[Rahimi and Recht, 2007]{rahimi2007random}
Rahimi, A. and Recht, B. (2007).
\newblock Random features for large-scale kernel machines.
\newblock {\em Advances in Neural Information Processing Systems}, 20.

\bibitem[Rasmussen and Williams, 2006]{gpml}
Rasmussen, C.~E. and Williams, C. K.~I. (2006).
\newblock {\em Gaussian Processes for Machine Learning}.
\newblock The MIT Press.

\bibitem[Roberts et~al., 2013]{timeseries-gp}
Roberts, S., Osborne, M., Ebden, M., Reece, S., Gibson, N., and Aigrain, S. (2013).
\newblock Gaussian processes for time-series modelling.
\newblock {\em Philosophical Transactions of the Royal Society A: Mathematical, Physical and Engineering Sciences}, 371(1984):20110550.

\bibitem[Rudin, 2017]{bochners}
Rudin, W. (2017).
\newblock {\em Fourier analysis on groups}.
\newblock Courier Dover Publications.

\bibitem[Rudner et~al., 2020]{iddgp2020}
Rudner, T.~G., Sejdinovic, D., and Gal, Y. (2020).
\newblock Inter-domain deep gaussian processes.
\newblock In {\em International Conference on Machine Learning}, pages 8286--8294. PMLR.

\bibitem[Salimbeni and Deisenroth, 2017]{dgp2017}
Salimbeni, H. and Deisenroth, M. (2017).
\newblock Doubly stochastic variational inference for deep gaussian processes.
\newblock {\em Advances in Neural Information Processing Systems}, 30.

\bibitem[Salimbeni et~al., 2019]{dgp2019importance}
Salimbeni, H., Dutordoir, V., Hensman, J., and Deisenroth, M. (2019).
\newblock Deep gaussian processes with importance-weighted variational inference.
\newblock In {\em International Conference on Machine Learning}, pages 5589--5598. PMLR.

\bibitem[Shen et~al., 2019]{pmlr-v89-shen19c}
Shen, Z., Heinonen, M., and Kaski, S. (2019).
\newblock Harmonizable mixture kernels with variational fourier features.
\newblock In Chaudhuri, K. and Sugiyama, M., editors, {\em Proceedings of the Twenty-Second International Conference on Artificial Intelligence and Statistics}, volume~89 of {\em Proceedings of Machine Learning Research}, pages 3273--3282. PMLR.

\bibitem[Snoek et~al., 2012]{bo2012}
Snoek, J., Larochelle, H., and Adams, R.~P. (2012).
\newblock Practical bayesian optimization of machine learning algorithms.
\newblock {\em Advances in neural information processing systems}, 25.

\bibitem[Titsias, 2009]{svgp2009}
Titsias, M. (2009).
\newblock Variational learning of inducing variables in sparse gaussian processes.
\newblock In van Dyk, D. and Welling, M., editors, {\em Proceedings of the Twelfth International Conference on Artificial Intelligence and Statistics}, volume~5 of {\em Proceedings of Machine Learning Research}, pages 567--574, Hilton Clearwater Beach Resort, Clearwater Beach, Florida USA. PMLR.

\bibitem[Van~der Wilk et~al., 2020]{van2020framework}
Van~der Wilk, M., Dutordoir, V., John, S., Artemev, A., Adam, V., and Hensman, J. (2020).
\newblock A framework for interdomain and multioutput gaussian processes.
\newblock {\em arXiv preprint arXiv:2003.01115}.

\bibitem[Wang et~al., 2018]{var-starvation2018}
Wang, Z., Gehring, C., Kohli, P., and Jegelka, S. (2018).
\newblock Batched large-scale bayesian optimization in high-dimensional spaces.
\newblock In {\em International Conference on Artificial Intelligence and Statistics}, pages 745--754. PMLR.

\end{thebibliography}

\appendix
\newpage
\onecolumn
\section*{Appendix}

\section{Model Comparison}
\label{app:model-comparison}

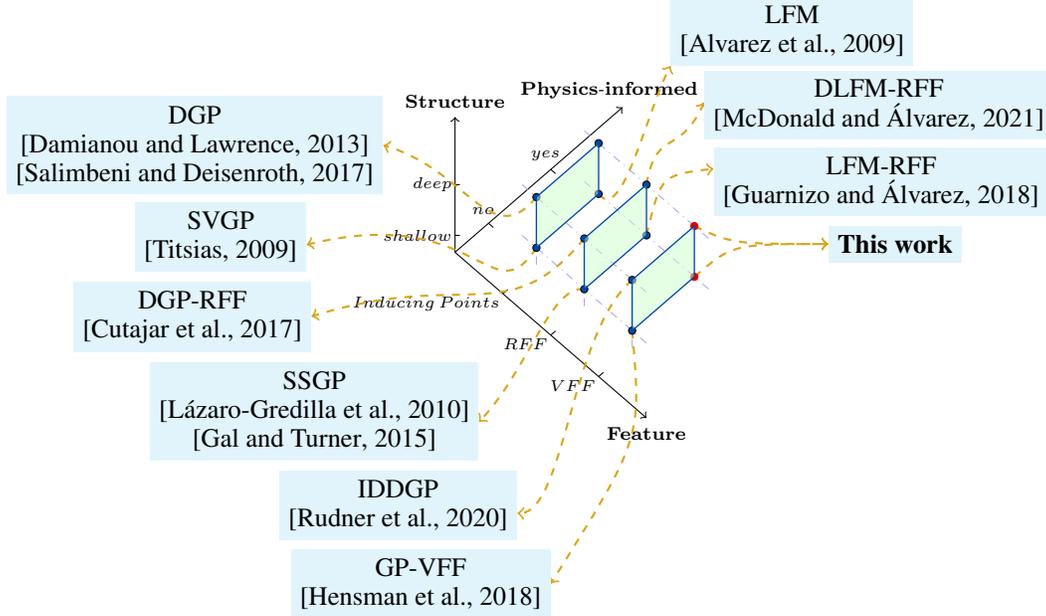
\begin{figure}[h]
\centering    
\begin{tikzpicture}[tdplot_main_coords,scale=1]
    \definecolor{myblue}{RGB}{0, 71, 171}
    \definecolor{myorange}{RGB}{221, 0, 0}
    \definecolor{myyellow}{RGB}{218, 165, 32}
    \definecolor{mylightblue}{RGB}{169, 169, 239}
    \definecolor{mypurple}{RGB}{220, 223, 240}

    \draw[thin,->] (0,0,0) -- (4.,0,0) node[anchor=north, font=\scriptsize] {$\mbf{Feature}$};
    \draw[thin,->] (0,0,0) -- (0,3.5,0) node[anchor=south, font=\scriptsize, xshift=-5pt]{$\mbf{Physics}$-$\mbf{informed}$};
    \draw[thin,->] (0,0,0) -- (0,0,4) node[anchor=south, font=\scriptsize]{$\mbf{Structure}$};
    
    \draw (1,0,0) -- (1,0.1,0) node[anchor=north east, font=\tiny, inner sep=1pt] at (1,0,0) {$Inducing\ Points$};
    \draw (2,0,0) -- (2,0.1,0) node[anchor=north east, font=\tiny, inner sep=1pt] at (2,0,0) {$RFF$};
    \draw (3,0,0) -- (3,0.1,0) node[anchor=north east, font=\tiny, inner sep=1pt] at (3,0,0) {$VFF$};
    \draw (0,0.7,0) -- (0.1,0.7,0) node[anchor=south, font=\tiny, xshift=-2pt] at (0,0.7,0) {$no$};
    \draw (0,2,0) -- (0.1,2,0) node[anchor=south, font=\tiny, xshift=-2pt] at (0,2,0) {$yes$};
    \draw (0,0,0.5) -- (0.05,0.05,0.5) node[anchor=east, font=\tiny, inner sep=1pt] at (0,0,0.5) {$shallow$};
    \draw (0,0,2) -- (0.05,0.05,2) node[anchor=east, font=\tiny, inner sep=1pt] at (0,0,2) {$deep$};

    \draw[dash dot,mylightblue] (0.7,0.7,0.5) -- (3.3,0.7,0.5);
    \draw[dash dot,mylightblue] (0.7,2,0.5) -- (3.3,2,0.5);
    \draw[dash dot,mylightblue] (0.7,0.7,2) -- (3.3,0.7,2);
    \draw[dash dot,mylightblue] (0.7,2,2) -- (3.3,2,2);
    \draw[dash dot,mylightblue] (1,0.7,0)--(1,0.7,2);
    \draw[dash dot,mylightblue] (2,0.7,0)--(2,0.7,2);
    \draw[dash dot,mylightblue] (3,0.7,0)--(3,0.7,2);

    \draw[thick, dashed, myyellow, ->] (1, 0.7, 0.5) .. controls (0.75, 0, -1) and (0.75, -0.25, 2.7) .. (0.4, -3.5, 5);
    \node[fill=cyan!10, anchor=east, align=center, yshift=3pt] at (0.4,-3.5,5) {SVGP \\ \citep{svgp2009}};
    \draw[thick, dashed, myyellow, ->] (1, 2, 0.5) .. controls (1, 2.5, -1) and (1, 3., 1.7) .. (0, 4.5, 0);
    \node[fill=cyan!10, anchor=south west, align=center] at (0,4.5,0) {LFM \\ \citep{lfms}};
    \draw[thick, dashed, myyellow, ->] (1, 0.7, 2) .. controls (0.75, 0, 0.9) and (0.75, -0.5, 4) .. (0., -1.5, 5);
    \node[fill=cyan!10, anchor=east, align=center] at (0,-1.5,5) {DGP \\ \citep{dgp2013} \\ \citep{dgp2017}};
    \draw[thick, dashed, myyellow, ->] (2, 0.7, 0.5) .. controls (2, 0, 1.5) and (2, -0.5, -0.5) .. (2.5, -2, 0.5);
    \node[fill=cyan!10, anchor=east, align=center, yshift=4pt] at (2.5,-2,0.5) 
    {SSGP\\ \citep{lazaro2010sparsespectrum} \\ \citep{gal2015improving}};
    \draw[thick, dashed, myyellow, ->] (2, 0.7, 2) .. controls (1.1, 0, -1) and (0.7, -0.5, 1) .. (0.5, -3.5, 3);
    \node[fill=cyan!10, anchor=east, align=center, yshift=0pt] at (0.5,-3.5,3) {DGP-RFF\\ \citep{cutajar2017random}};
    \draw[thick, dashed, myyellow, ->] (2, 2, 2) .. controls (1., 3.2, 1) and (1.2,3.5,0) .. (1.2, 4, 1);
    \node[fill=cyan!10, anchor=west, align=center, yshift=-1pt] at (1.2,4,1) {DLFM-RFF\\ \citep{dlfm2021}};
    \draw[thick, dashed, myyellow, ->] (2, 2, 0.5) .. controls (1.3,3,0.2) .. (2.1, 3.3, 1);
    \node[fill=cyan!10, anchor=west, align=center, yshift=-5pt] at (2.1,3.3,1) {LFM-RFF\\ \citep{fastkernel}};
    \draw[thick, dashed, myyellow, ->] (3, 0.7, 2) .. controls (3.4, -1, 3) and (3.5, -1.2, -2) ..  (3.8, -2.5, 0);
    \node[fill=cyan!10, anchor=east, align=center, yshift=3pt] at (3.8,-2.5,0) {IDDGP\\ \citep{iddgp2020}};
    \draw[thick, dashed, myyellow, ->] (3, 0.7, 0.5) .. controls (4.2, -0.3, 0) ..  (5, -3, 0);
    \node[fill=cyan!10, anchor=east, align=center, yshift=0pt] at (5,-3,0) {GP-VFF\\ \citep{vffs}};
    \draw[thick, dashed, myyellow, ->] (3, 2, 2) .. controls (2.8,3,0) .. (3.8, 4, 0);
    \node[fill=cyan!10, anchor=west, align=center] at (3.8,4,0) {\textbf{This work}};
    \draw[thick, dashed, myyellow, ->] (3, 2, 0.5) .. controls (2.8,3,0) .. (3.8, 4, 0);

    \draw[fill=myblue] (1,0.7,0.5) circle (1.5pt);
    \draw[fill=myblue] (1,2,0.5) circle (1.5pt);
    \draw[fill=myblue] (1,0.7,2) circle (1.5pt);
    \draw[fill=myblue] (1,2,2) circle (1.5pt);
    \filldraw[fill=green!20, fill opacity=0.5, draw=myblue, line width=0.5pt] (1,0.7,0.5) -- (1,2,0.5) -- (1,2,2) -- (1,0.7,2) -- cycle;

    \draw[fill=myblue] (2,0.7,0.5) circle (1.5pt);
    \draw[fill=myblue] (2,2,0.5) circle (1.5pt);
    \draw[fill=myblue] (2,0.7,2) circle (1.5pt);
    \draw[fill=myblue] (2,2,2) circle (1.5pt);
    \filldraw[fill=green!20, fill opacity=0.5, draw=myblue, line width=0.5pt] (2,0.7,0.5) -- (2,2,0.5) -- (2,2,2) -- (2,0.7,2) -- cycle;

    \draw[fill=myblue] (3,0.7,0.5) circle (1.5pt);
    \draw[fill=myorange, draw=myorange] (3,2,0.5) circle (1.5pt);
    \draw[fill=myblue] (3,0.7,2) circle (1.5pt);
    \draw[fill=myorange, draw=myorange] (3,2,2) circle (1.5pt);
    \filldraw[fill=green!20, fill opacity=0.5, draw=myblue, line width=0.5pt] (3,0.7,0.5) -- (3,2,0.5) -- (3,2,2) -- (3,0.7,2) -- cycle;
\end{tikzpicture}
\caption{An illustration on a comparison of our model with recent related work. The comparison dimensions are the used feature, the structure depth, and whether incorporating physical dynamics.}
\label{fig:model-comparison}
\end{figure}

We give a high-level comparison of our approach with other related models in Fig.~\ref{fig:model-comparison}. Each point represents a corresponding model. The reference is attached to the blue tag. The comparison dimensions include whether the model has a multi-layer structure, whether it incorporates physics-informed modelling involving ODEs and convolutions, and which kind of feature it uses. The red points represent our work.
\tdplotsetmaincoords{30}{45}  

\section{Variational Fourier Feature}
\label{app:vff}

Variational Fourier Features (VFFs, \citep{vffs}) are designed on a \matern\ Reproducing Kernel Hilbert Space (RKHS). Specifically, \matern-$\frac{1}{2}/\frac{3}{2}/\frac{5}{2}$ kernels with inputs $x, x'\in\RR $ are of particular interest:
\begin{gather}
    k_{1/2}(r) = \sigma^2 e^{-\lambda r}, \ \lambda = \frac{1}{l}, \\
    k_{3/2}(r) = \sigma^2(1+\lambda r)e^{-\lambda r},\ \lambda=\frac{\sqrt{3}}{l}, \\
    k_{5/2}(r) = \sigma^2(1+\lambda r +\frac{1}{3}\lambda^2 r^2)e^{-\lambda r},\ \lambda=\frac{\sqrt{5}}{l},
\end{gather}
where $r=|x - x'|$, $\sigma^2$ is the kernel's output-scale (or variance), and $l$ is the length-scale. We reiterate the closed-form RKHS inner products for \matern-$\frac{1}{2}$ on $[a, b]$ here ( for other \matern\ kernels and more details see \citet{durrande2016detecting}) : 
\begin{equation}
    \langle g, h\rangle_{\mathcal{H}_{\frac{1}{2}}}=
    \frac{1}{2\lambda\sigma^2}
    \int_{a}^b (\lambda g(x)+g'(x))(\lambda h(x)+h'(x))\dif x + \frac{1}{\sigma^2}g(a)h(a).
\end{equation}
The explicit expressions of the \matern\ RKHS not only allow us to verify the reproducing property when $t\in[a,b]$
\begin{equation}
    \langle k(t, \cdot), h(\cdot)\rangle_{\mtc{H}}= h(t), \ \forall h\in\mtc{H}, t\in[a,b],
\end{equation}
 but also make it feasible to complete the cross-covariance if $t$ is outside $[a, b]$. In Appx.~\ref{app:mkg}, we further utilise the conclusion of VFFs in \cite{vffs} for $t\in\RR$ to calculate our \emph{Variational Fourier Response Features} (VFRFs). 

Given a \matern\ GP $f(t)\sim\mtc{GP}(0, k(t, t'))$, the explicit RKHS inner product provides an alternative linear operator to construct an inter-domain GP by $
    u_m= \langle f, \phi_m\rangle_{\mathcal{H}}$, where $\phi_m, m=0,\ldots,2M$ is from a set of truncated Fourier basis with harmonic \emph{inducing frequencies}
\begin{equation}
    \phi_0(\cdot) = 1,\ \phi_m(\cdot)=\cos(z_m(\cdot-a)),\ \phi_{M+m}(\cdot)=\sin(z_m(\cdot-a)),\ z_m=\frac{2\pi m}{b-a}.
\end{equation}
$u(\cdot)$ is an \emph{inter-domain} GP sharing a joint Gaussian prior with $f(\cdot)$. For $t\in[a,b]$, the covariances are
\begin{equation}
    \mathrm{Cov}[f(t), u_m]=\langle k(t,\cdot), \phi_m(\cdot)\rangle_{\mtc{H}}=\phi_m(t),\quad
    \mathrm{Cov}[u_i, u_j]=\langle\phi_i(\cdot),\phi_j(\cdot)\rangle_{\mtc{H}}.
\end{equation}
The VFFs approximate the posterior by replacing the covariance matrix in Sparse Variational GPs (SVGPs) appropriately.

\section{LFMs for First-Order Dynamical System}
\label{app:lfm-kernel}

We recall in this work a dynamical system modelled by a first-order ODE 
\begin{equation}
    \beta \frac{\dif f(t)}{\dif t} + \alpha f(t) = u(t),\ u(t)\sim\mtc{GP}(0, k(t,t')),
    \ \alpha, \beta>0,    
    \label{eq:ode-1}
\end{equation}
where $u$ is an unobserved latent force with a \matern\ kernel. The Green's function
is $G(t)=\frac{1}{\beta}\exp\left(-\gamma t\right), \gamma=\frac{\alpha}{\beta}$. We take the solution
\begin{equation}
    f(t) = \int_{-\infty}^t G(t-\tau)u(\tau)\dif\tau=G\circ u,
\end{equation}
where a convolutional operator ${G}$ acting on $u$ is represented as $f = G\circ u$. Conventional LFMs establish a GP over $f\sim\mtc{GP}(0, G\circ k\circ G)$, where $k\circ G$ signifies $G$ operating the second argument of the kernel. The LFM kernels are computed analytically in \cite{lawrence2006modelling, lfms}, but their expressions are based on RBF kernels. In the subsequent part, we will present the closed-form covariance expressions $G\circ k\circ G$ for the \matern-$\frac{1}{2}/\frac{3}{2}/\frac{5}{2} $ kernels, respectively. All analytical LFM covariance functions discussed in this work are summarised in Table~\ref{tab:gkg} and illustrated in the left panel of Fig.~\ref{fig:basis-functions}, \ref{fig:basis-function32} and \ref{fig:basis-function52}. Furthermore, we introduce the approximation of our LFM kernels using random Fourier features in Appx.~\ref{app:fastkernel}.

\subsection{Analytical LFM \texorpdfstring{\matern\ }{Matern} Kernels}

The LFM kernel of a \matern-$\frac{1}{2}$ latent force when $t>t'$ is given by
\begin{align}
    G\circ k\circ G 
    & = 
    \int_{-\infty}^t\int_{-\infty}^{t'}\frac{1}{\beta}e^{-\gamma(t-\tau)}
    \cdot \sigma^2 e^{-\lambda|\tau-\tau'|} \cdot \frac{1}{\beta}e^{-\gamma(t'-\tau')}\dif\tau\dif\tau' \notag \\
    & =
    \frac{\sigma^2}{\beta^2}\int_{-\infty}^{t'}
    \int_{\tau'}^{t}e^{-\gamma(t-\tau) -\lambda(\tau-\tau') -\gamma(t'-\tau')}\dif\tau \dif\tau'   \notag\\
    & \quad + \frac{\sigma^2}{\beta^2}\int_{-\infty}^{t'}
    \int_{-\infty}^{\tau'}e^{-\gamma(t-\tau) - \lambda(\tau'-\tau) -\gamma(t'-\tau')}\dif\tau \dif\tau' \notag \\
    & = 
    \frac{\sigma^2}{\beta^2\gamma(\gamma^2-\lambda^2)}\left[\gamma e^{-\lambda(t-t')}- \lambda e^{-\gamma(t-t')}\right].
\end{align}

The derivation for $t<t'$ is similar. As a result, we obtain a stationary LFM kernel for $\forall t, t'\in\RR$,
\begin{align}
    G\circ k\circ G =
    \begin{cases}
        \frac{\sigma^2}{\beta^2\gamma(\gamma^2 - \lambda^2)}\left[\gamma e^{-\lambda|t-t'|}- \lambda e^{-\gamma|t-t'|}\right]
        & \text{if}\ \gamma\neq \lambda, \\
        \addls
        \frac{\sigma^2 (1+\lambda|t-t'|)}{2\beta^2\lambda^2} e^{-\lambda|t-t'|}
        & \text{if}\ \gamma = \lambda.
    \end{cases}
\end{align}
Likewise, the expressions of the other LFM kernels are present in Table~\ref{tab:gkg}. The expressions exhibit continuity but non-differentiability at the point where $\gamma=\lambda$.

\begin{table}[thb]
  \centering
  \begin{threeparttable}
    \caption{LFM kernels of \matern-$\tfrac{1}{2}/\tfrac{3}{2}/\tfrac{5}{2}$ latent forces}    
    \begin{tabular}{cl}
    \toprule
    Latent Force $k(r)$  & LFM $G\circ k\circ G$ \tnote{1} \\
    \midrule
    \matern-$\frac{1}{2}$ 
    & $\frac{\sigma^2}{\beta^2\gamma(\gamma^2 - \lambda^2)}\left[\gamma e^{-\lambda r}- \lambda e^{-\gamma r}\right]$ \\ 
    \addls
    \matern-$\frac{3}{2}$ 
    & $\frac{\sigma^2}{\beta^2}\left[\frac{\lambda r+1}{\gamma^2-\lambda^2}-\frac{2\lambda^2}{(\gamma^2-\lambda^2)^2}\right]e^{-\lambda r} + \frac{2\lambda^3\sigma^2}{\beta^2\gamma(\gamma^2-\lambda^2)^2}e^{-\gamma r}$\\
    \addls
    \matern-$\frac{5}{2}$ & $\frac{\sigma^2}{3\beta^2}\left[
        \frac{\lambda^2 r^2+3}{\gamma^2-\lambda^2}+\frac{\lambda(3\gamma^2-7\lambda^2)r}{(\gamma^2-\lambda^2)^2}
        +\frac{4\lambda^2(3\lambda^2-\gamma^2)}{(\gamma^2-\lambda^2)^3}\right]e^{-\lambda r}
         - \frac{8\lambda^5\sigma^2}{3\beta^2\gamma(\gamma^2-\lambda^2)^3}e^{-\gamma r}$ \\
    \bottomrule
    \end{tabular}
    \label{tab:gkg}  
    \begin{tablenotes}
      \item[1] $r$ is the input distance $r=|t-t'|$.  $\{\beta, \gamma\}$ and $\{\sigma^2, \lambda\}$ are parameters from the ODE and the \matern\ kernel, respectively. 
    \end{tablenotes}
  \end{threeparttable}
\end{table}

\subsection{Random Fourier Approximation of Kernels}
\label{app:fastkernel}

Stationary kernels can be approximated using random Fourier features \citep{rahimi2007random} using Bochner's theorem \citep{bochners}. In the illustrative experiment, we approximate the \matern\ LFM covariance proposed in our work using random Fourier features (also termed \emph{Random Fourier Response Features} (RFRFs) by \citep{fastkernel}). The random Fourier features of the LFM with a \matern-$\nu\ (\nu=\tfrac{1}{2}/\tfrac{3}{2}/\tfrac{5}{2})$ kernel is given by
\begin{equation}
    \varphi(t; \omega) = \int_{-\infty}^t e^{j\omega\tau}\cdot\frac{1}{\beta}e^{-\gamma (t-\tau)}\dif\tau = \frac{e^{j\omega t}}{\beta(\gamma+j\omega)},\ 
    \omega=\frac{\omega'}{l},\ \omega'\sim t_{2\nu}(\omega'),
\end{equation}
where $j^2=-1$, $t_{2\nu}$ is a zero-mean Student's $t$-distribution with $2\nu$ degrees of freedom, and $l$ is the length-scale of the \matern\ kernel. Therefore,
\begin{align}
    G\circ k & \approx \frac{\sigma^2}{M}\sum_{m=1}^{M}\varphi(t; \omega_m)\cdot 
    e^{-j\omega_m t'},\\
    G\circ k\circ G &\approx \frac{\sigma^2
    }{M}\sum_{m=1}^{M}\varphi(t; \omega_m)\cdot \Bar{\varphi}(t';\omega_m),
\end{align}
where $\sigma^2$ is \matern\ kernel's variance, $\{\omega_m\}_{m=1}^M$ are $M$ random Fourier frequencies sampled from the corresponding Student's $t$-distribution. $\Bar{\varphi}$ denotes the complex conjugate of $\varphi$.

\section{Variational Fourier Response Features for LFMs}
\label{app:mkg}

We represent the projection of the latent force $u$ onto the truncated Fourier basis $\phi$ as
\begin{equation}
    v(z) = P \circ u = \langle \phi(\cdot; z),  u(\cdot)\rangle_{\mtc{H}},
\end{equation}
where $v(z)$ is the projection process for the latent force, and $z$ is the inducing frequency. For simplicity, the projection operator is denoted as $P$. Consequently, the output process $f$ and the projection process $v$ share a joint GP prior: 
\begin{equation}
\left[\begin{array}{c}
    f \\ v
\end{array}\right]
\sim\mathcal{N}
\left(\mbf{0},
\left[\begin{array}{cc}
    G\circ k\circ G & G \circ k\circ P  \\
    P\circ k\circ G & P \circ k\circ P
\end{array}\right]
\right),
\label{whole-prior}
\end{equation}
where the covariance terms are given by
\begin{align}
    \mathrm{Cov}[f(t), v(z)]
    &=
    \mathbb{E}\left[f(t)\langle u(\cdot),\phi(\cdot; z)\rangle_{\mtc{H}}\right]
    = \langle \mathbb{E}[f(t)u(\cdot)],\phi(\cdot; z) \rangle_{\mtc{H}} \notag \\
    &=
    \langle G\circ k(t,\cdot), \phi(\cdot; z) \rangle_{\mtc{H}}
    = G\circ \langle k(t,\cdot), \phi(\cdot; z)\rangle_{\mathcal{H}}
    = G\circ k\circ P. \\
    \mathrm{Cov}[v(z_i),v(z_j)] & = \langle\phi(\cdot;z_i),\phi(\cdot;z_j)\rangle_{\mtc{H}} = P\circ k\circ P.
\end{align}
The rest of this section will specify the closed-form VFRF expressions of $P\circ k\circ G$. With the integration lower limit going to negative infinity, the input values outside the interval $[a,b]$ should be considered. 

\begin{figure}[htbp]
    \centering
    \includegraphics[width=1\linewidth]{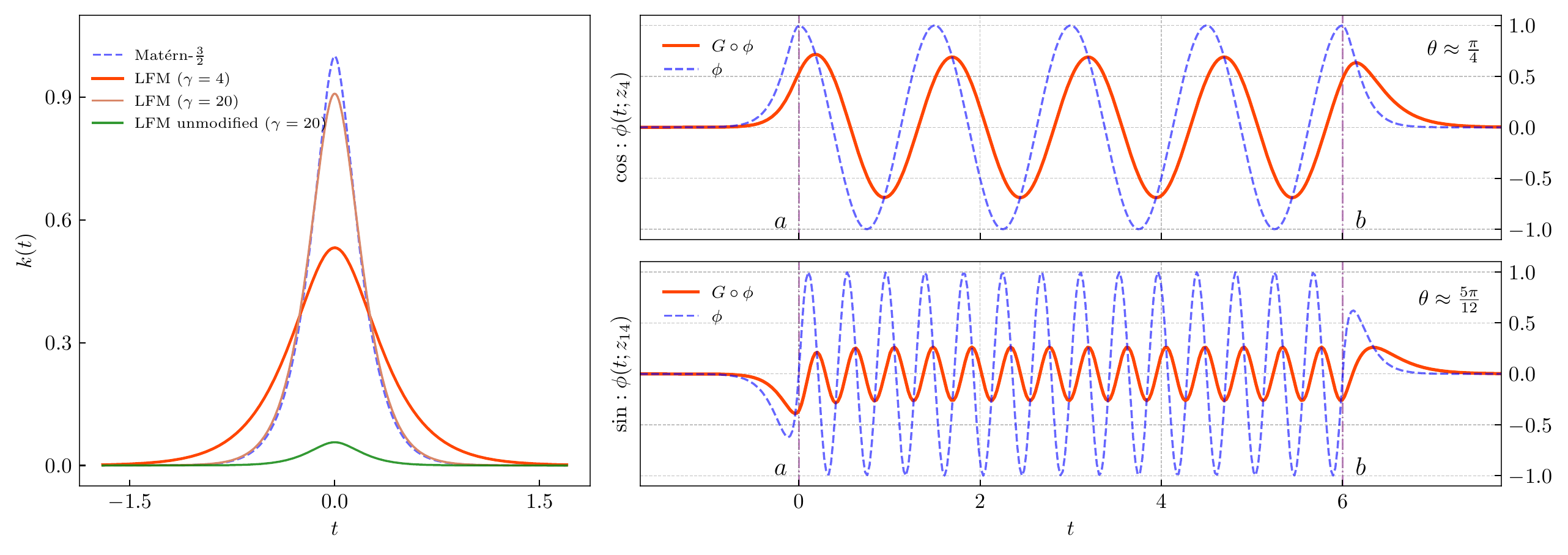}
    \caption{Covariance functions and VFRFs for LFMs with \matern-$\frac{3}{2}$ kernel.}
    \label{fig:basis-function32}
\end{figure}

\begin{figure}[ht]
    \centering
    \includegraphics[width=1\linewidth]{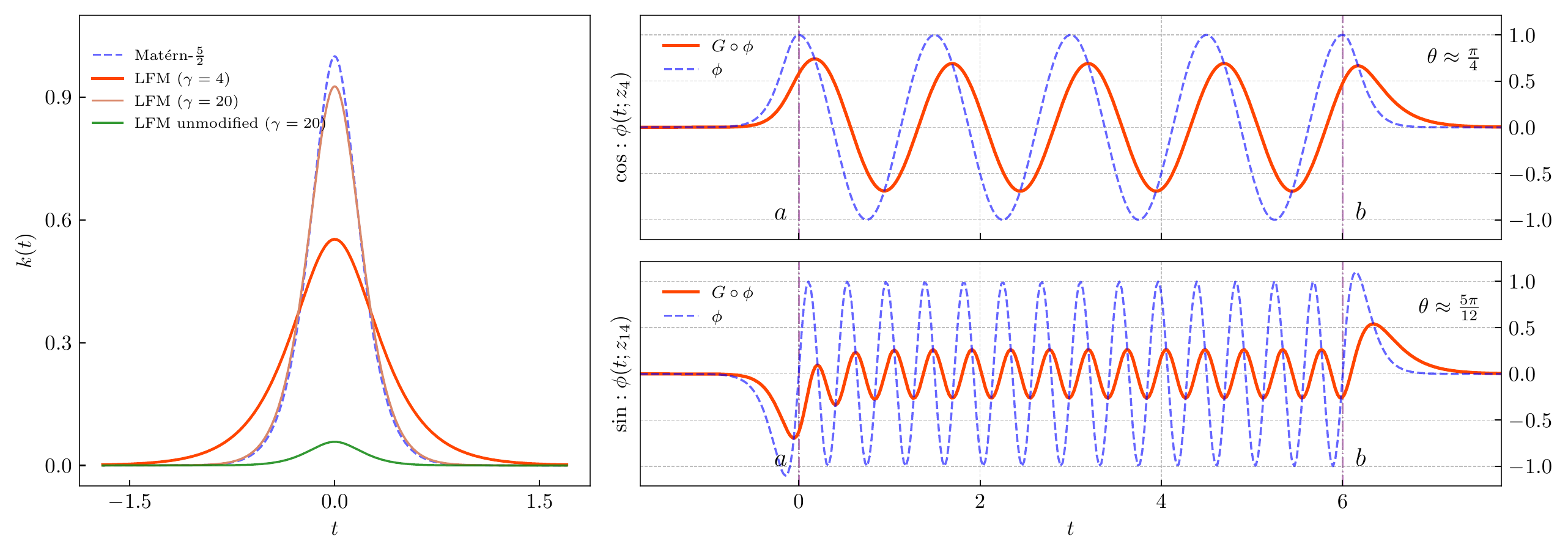}
    \caption{Covariance functions and VFRFs for LFMs with \matern-$\frac{5}{2}$ kernel.}
    \label{fig:basis-function52}
\end{figure}

We detail the derivation of \matern-$\frac{1}{2}$ (listed in Table~\ref{tab:mkg-12-cos} and \ref{tab:mkg-12-sin}) and directly give the results of \matern-$\frac{3}{2}$ (see Table~\ref{tab:mkg-32-cos} and \ref{tab:mkg-32-sin}) and \matern-$\frac{5}{2}$ (see Table~\ref{tab:mkg-52-cos} and \ref{tab:mkg-52-sin}). These expressions of VFRFs will revert to VFFs under certain conditions. The LFM kernel and the VFRFs for \matern-$\frac{3}{2}$ and \matern-$\frac{5}{2}$ with the same hyperparameters in the main text are depicted in Fig.~\ref{fig:basis-function32} and Fig.~\ref{fig:basis-function52}, respectively.

\subsection{\texorpdfstring{\matern-$\frac{1}{2}$}{Matern-1/2} Cross-covariance}

We write the cross-covariance as 
\begin{equation}
    P\circ k \circ G 
    = \langle\phi(\cdot;z),k(\cdot, \tau')\rangle_{\mtc{H}}\circ G
    = \int_{-\infty}^{t'} h(z,\tau')\cdot G(t'-\tau')\dif\tau'.
    \label{eq:mkg-12}
\end{equation}
The location of $\tau$ determines the expression of $h(z, \tau')=\langle\phi(\cdot;z),k(\cdot, \tau')\rangle_{\mtc{H}}$ (i.e., $P\circ k$) \citep{vffs}, as summarized in the subsequent table (For \matern-$\frac{3}{2}/\frac{5}{2}$, see Table~\ref{tab:hzx-32}/\ref{tab:hzx-52}). The columns of the tables indicate the VFFs for input $\tau'$ located inside/outside $[a,b]$. The cross-covariance can be derived by substituting $h(z, \tau')$ in \eqref{eq:mkg-12}. Table~\ref{tab:mkg-12-cos} and Table~\ref{tab:mkg-12-sin} collect the VFRFs for \matern-$\frac{1}{2}$ LFMs with cosine/sine projection basis functions at different locations of $t'$ in \eqref{eq:mkg-12}.
\begin{table}[htb]
    \centering
    \caption{VFFs $h(z, \tau')$ with \matern-$\frac{1}{2}$ kernel}
    \begin{tabular}{cccc}
    \toprule
     $\phi(\cdot; z)$& $\tau'<a$& $a \le  \tau'\le b$& $\tau'>b$\\
     \midrule
     $\cos(z(\cdot-a))$& $e^{-\lambda(a-\tau')}$& $\cos(z(\tau'-a))$& $e^{-\lambda(\tau'-b)}$\\
     \addls
     $\sin(z(\cdot-a))$& 0                      & $\sin(z(\tau'-a))$& 0                     \\
     \bottomrule
    \end{tabular}
    \label{tab:hzx-12}
\end{table} 
\subsubsection{Cosine Features (\texorpdfstring{$\phi(t; z)=\cos(z(t-a))$}{phi=cos(z(t-a)})}

\paragraph{Case 1:} $t'< a$,
\begin{equation}
    P\circ k\circ G =
    \int_{-\infty}^{t'}e^{-\lambda(a-\tau')}\cdot\frac{1}{\beta}e^{-\gamma(t'-\tau')}\dif\tau' =
    \frac{1}{\beta(\gamma+\lambda)}e^{-\lambda(a-t')}.
\end{equation}
The covariance will converge to the VFF in Table~\ref{tab:hzx-12} with a scaling coefficient $\alpha$:
\begin{equation}
    \lim\limits_{\beta\to 0^{+}} P\circ k\circ G = \frac{1}{\alpha}e^{-\lambda(a-t')}.
\end{equation}

\paragraph{Case 2:} $a\leq t'\leq b$,
\begin{align}
    P\circ k \circ G
    & =
    \int_{-\infty}^{a}e^{-\lambda(a-\tau')}\cdot\frac{1}{\beta}e^{-\gamma(t'-\tau')}\dif\tau' 
    + \int_a^{t'} \cos(z(\tau'-a))\cdot\frac{1}{\beta}e^{-\gamma(t'-\tau')}\dif\tau' \notag \\
    &=
    \frac{\gamma\cos(z (t'-a))+z\sin(z (t'-a))}{\beta(z^2+\gamma^2)} 
    + \frac{(z^2-\gamma\lambda)}{\beta(\gamma+\lambda)(z^2+\gamma^2)}e^{-\gamma(t'-a)} \notag\\
    &=
    \frac{\cos(z(t'-a)+\theta)}{\beta\sqrt{z^2+\gamma^2}} + \xi_{\cos},
\end{align}
where $\theta=-\arctan(\frac{z}{\gamma})$, and $\xi$ is a decay term. Particularly, the cross-covariance extends the VFFs of the latent force since
\begin{equation}
    \lim\limits_{\beta\to 0_{+}}P\circ k\circ G = \frac{1}{\alpha}\cos(z(t'-a)),
\end{equation}
    
\paragraph{Case 3:} $t'> b$,
\begin{equation}
    P\circ k \circ G =
    \frac{e^{-\lambda(t'-b)}}{\beta(\gamma-\lambda)} 
    + \frac{(z^2-\gamma\lambda)e^{-\gamma(t'-a)}}{\beta(\gamma+\lambda)(z^2+\gamma^2)}
    -\frac{(z^2+\gamma\lambda)e^{-\gamma(t'-b)}}{\beta(\gamma-\lambda)(z^2+\gamma^2)},
\end{equation}

which utilizes harmonic $z=\frac{2\pi m}{b - a}, m\in\mathbb{Z}_+$.  The covariance will also return to a scaled term in Table~\ref{tab:hzx-12} when $\beta\rightarrow 0_{+}$.

\subsubsection{Sine Features (\texorpdfstring{$\phi(t;z)=\sin(z(t-a))$}{phi=sin(z(t-a)})}

\paragraph{Case 1:}  $t'< a,\ P\circ k\circ G = 0. $

\paragraph{Case 2:} $a \leq t'\leq b$,
\begin{align}
    P\circ k \circ G 
    & =
    \int_a^{t'} \sin(z(\tau'-a))\cdot\frac{1}{\beta}e^{-\gamma(t'-\tau')}\dif\tau' \notag \\ 
    & = 
    \frac{- z\cos(z (t'-a)) + \gamma\sin(z (t'-a))}{\beta(z^2+\gamma^2)} + \frac{z}{\beta(z^2+\gamma^2)}e^{-\gamma(t'-a)} \notag\\
    & =
    \frac{\sin(z(t'-a)+\theta)}{\beta\sqrt{z^2+\gamma^2}} + \xi_{\sin},
    \quad \theta=-\arctan(\frac{z}{\gamma})
\end{align}

\paragraph{Case 3:} $t' > b$,
\begin{equation}
    P\circ k\circ G =
    \int_a^{b} \sin(z(\tau'-a))\cdot\frac{1}{\beta}e^{-\gamma(t'-\tau')}\dif\tau' =
    \frac{z e^{-\gamma(t'-a)}- z e^{-\gamma(t'-b)}}{\beta(\gamma^2+z^2)}.
\end{equation}

The VFRFs of the LFM with a \matern-$\frac{1}{2}$ latent force are summarised in Table~\ref{tab:mkg-12-cos} and \ref{tab:mkg-12-sin}, where the absolute distances to the interval ends are denoted as $r_a=|t'-a|$ and $r_b=|t'-b|$ and the phase shift is $\theta=-\arctan(\frac{z}{\gamma})$. The features are continuous at $\gamma=\lambda$ when $t'>b$.
\begin{table}[htbp]
    \centering
    \caption{\matern-$\tfrac{1}{2}$ VFRFs on Fourier basis $\phi(x;z)=\cos(z(x-a))$. }
    \begin{tabular}{cl}
    \toprule
    $t'\in\RR$ &  LFM Fourier Feature $P\circ k\circ G$ (cosine part)\\
    \midrule
    $t'< a $ & $\frac{1}{\beta(\gamma+\lambda)}e^{-\lambda r_a}$  \\
    \addls
    $a\le t'\le b$  
    & $ \frac{\cos(z r_a +\theta)}{\beta\sqrt{z^2+\gamma^2}}
        - \left[\frac{\gamma}{\beta(z^2+\gamma^2)}-\frac{1}{\beta(\gamma+\lambda)}\right]e^{-\gamma r_a}$ \\
    \addls
    $t'>b\ (\gamma\neq \lambda)$ 
    & $ - \left[\frac{\gamma}{\beta(z^2+\gamma^2)} - \frac{1}{\beta(\gamma+\lambda)}\right]e^{-\gamma r_a}
        + \left[\frac{\gamma}{\beta(z^2+\gamma^2)} - \frac{1}{\beta(\gamma-\lambda)}\right]e^{-\gamma r_b}
        + \frac{1}{\beta(\gamma-\lambda)}e^{-\lambda r_b}$ \\
    \addls
    $t'>b\ (\gamma=\lambda)$ 
    & $- \left[\frac{\lambda}{\beta(z^2+\lambda^2)}-\frac{1}{2\beta\lambda}\right]e^{-\lambda r_a} 
        + \left[\frac{\lambda}{\beta(z^2+\lambda^2)}+\frac{r_b}{\beta}\right]e^{-\lambda r_b}$ \\
    \bottomrule
    \end{tabular}
    \label{tab:mkg-12-cos}
\end{table}
\begin{table}[htbp]
    \centering   
    \caption{\matern-$\tfrac{1}{2}$ VFRFs on Fourier basis $\phi(x;z)=\sin(z(x-a))$. }
    \begin{tabular}{cl}
    \toprule
    $t'\in\RR$ & LFM Fourier Feature $P\circ k\circ G$ (sine part)\\
    \midrule
    $t'<a$  & \makecell[l]{$0$} \\
    \addls
    $a\leq t'\le b$  & $\frac{\sin(z r_a +\theta)}{\beta\sqrt{z^2+\gamma^2}}+ \frac{z}{\beta(z^2+\gamma^2)}e^{-\gamma r_a}$ \\
    \addls
    $t'>b$ & $\frac{z}{\beta(z^2+\gamma^2)}e^{-\gamma r_a} - \frac{z}{\beta(z^2+\gamma^2)}e^{-\gamma r_b}$ \\
    \bottomrule
    \end{tabular}
    \label{tab:mkg-12-sin}
\end{table}

\subsection{\texorpdfstring{\matern-$\frac{3}{2}/\frac{5}{2}$}{Matern-3/2-5/2} Cross-covariance}

Based on Table~\ref{tab:hzx-32} and \ref{tab:hzx-52}, we give the VFRFs for \matern-$\frac{3}{2}$ and \matern-$\frac{5}{2}$ LFMs with $\theta=-\arctan(\frac{z}{\gamma})$ in Table~\ref{tab:mkg-32-cos},\ref{tab:mkg-32-sin},\ref{tab:mkg-52-cos} and \ref{tab:mkg-52-sin}. Also, the derived cross-covariance expressions can return to the scaled VFFs of the latent force.

\begin{table}[htbp]
    \centering
    \caption{VFFs $h(z, \tau')$ with \matern-$\frac{3}{2}$ kernel}
    \begin{tabular}{cccc}
    \toprule
     $\phi(\cdot;z)$& $\tau'<a$& $a \le  \tau'\le b$& $\tau'>b$\\
     \midrule
     $\cos(z(\cdot-a))$& $(1+\lambda(a-\tau'))e^{-\lambda(a-\tau')}$& $\cos(z(\tau'-a))$& $(1+\lambda(\tau'-b))e^{-\lambda(\tau'-b)}$\\
     \addls
     $\sin(z(\cdot-a))$& $z(\tau'-a) e^{-\lambda(a-\tau')}$& $\sin(z(\tau'-a))$& $z(\tau'-b) e^{-\lambda(\tau'-b)}$\\
     \bottomrule
    \end{tabular}
    \label{tab:hzx-32}
\end{table} 

\begin{table}[htbp]
    \centering
    \caption{\matern-$\tfrac{3}{2}$ VFRFs on Fourier basis $\phi(x;z)=\cos(z(x-a))$. }
    \begin{tabular}{cl}
    \toprule
    $t'\in\RR$ & LFM Fourier Feature $P\circ k\circ G$ (cosine part)\\
    \midrule
    $t'<a$  & $ \left[\frac{\lambda r_a + 1}{\beta(\gamma+\lambda)}+\frac{\lambda}{\beta(\gamma+\lambda)^2}\right]e^{-\lambda r_a} $  \\
    \addls
    $a\leq t'\le b$  
    & $\frac{\cos(z r_a+\theta)}{\beta\sqrt{z^2 + \gamma^2}}
        - \left[\frac{\gamma}{\beta(z^2+\gamma^2)}-\frac{\gamma+2\lambda}{\beta(\gamma+\lambda)^2}\right]e^{-\gamma r_a} $ \\
    \addls
    $t'>b\ (\gamma\neq\lambda)$ 
    & \makecell[l]{
        $ - \left[\frac{\gamma}{\beta(z^2+\gamma^2)}-\frac{\gamma+2\lambda}{\beta(\gamma+\lambda)^2}\right]e^{-\gamma r_a}
        + \left[\frac{\gamma}{\beta(z^2+\gamma^2)}-\frac{\gamma-2\lambda}{\beta(\gamma-\lambda)^2} \right]e^{-\gamma r_b}$ \\
        \addlinespace[1pt]
        $ \quad + \left[\frac{\lambda r_b + 1}{\beta(\gamma-\lambda)}-\frac{\lambda}{\beta(\gamma-\lambda)^2}\right]e^{-\lambda r_b}$ 
        }\\
    \addls
    $t'>b\ (\gamma=\lambda)$ 
    & $-\left[\frac{\lambda}{\beta(z^2+\lambda^2)}-\frac{3}{4\beta\lambda}\right]e^{-\lambda r_a}
    + \left[\frac{\lambda}{\beta(z^2+\lambda^2)}+\frac{(\lambda r_b +2)r_b}{2\beta}\right]e^{-\lambda r_b}$\\
    \bottomrule
    \end{tabular}  
    \label{tab:mkg-32-cos}
\end{table}

\begin{table}[htbp]
    \centering
    \caption{\matern-$\tfrac{3}{2}$ VFRFs on Fourier basis $\phi(x;z)=\sin(z(x-a))$. }
    \begin{tabular}{cl}
    \toprule
    $t'\in\RR$ & LFM Fourier Feature $P\circ k\circ G$ (sine part)\\
    \midrule
    $t'<a$  &  $-\frac{z}{\beta}\left[\frac{r_a}{\gamma+\lambda}+\frac{1}{(\gamma+\lambda)^2}\right]e^{-\lambda r_a}  $ \\
    \addls
    $a\leq t'\le b$  
    & $\frac{\sin(z r_a+\theta)}{\beta\sqrt{z^2 + \gamma^2}}
        + \frac{z}{\beta}\left[\frac{1}{(z^2+\gamma^2)}-\frac{1}{(\gamma+\lambda)^2} \right]e^{-\gamma r_a} $ \\
    \addls
    $t'>b\ (\gamma\neq\lambda)$ 
    & \makecell[l]{
        $ \frac{z}{\beta}\left[\frac{1}{(z^2+\gamma^2)}-\frac{1}{(\gamma+\lambda)^2}\right]e^{-\gamma r_a}
        - \frac{z}{\beta}\left[\frac{1}{(z^2+\gamma^2)} - \frac{1}{(\gamma-\lambda)^2}\right]e^{-\gamma r_b}$ \\
        \addlinespace[1pt]
        $ \quad + \frac{z}{\beta}\left[\frac{r_b}{\gamma-\lambda}-\frac{1}{(\gamma-\lambda)^2}\right]e^{-\lambda r_b}$ 
        }\\
    \addls
    $t'>b\ (\gamma=\lambda)$ 
    & $ \frac{z}{\beta}\left[\frac{1}{(z^2+\lambda^2)}-\frac{1}{4\lambda^2}\right]e^{-\lambda r_a}
        + \frac{z}{\beta}\left[-\frac{1}{(z^2+\lambda^2)}+\frac{r_b^2}{2}\right]e^{-\lambda r_b}$ \\
    \bottomrule
    \end{tabular}
    \label{tab:mkg-32-sin}
\end{table}

\begin{table}[htbp]
\centering
\begin{threeparttable}
    \centering
    \caption{VFFs $h(z, \tau')$ with \matern-$\frac{5}{2}$ kernel \tnote{1}}
    \begin{tabular}{cccc}
    \toprule
     $\phi(\cdot,z)$& $\tau'<a$& $a \le  \tau'\le b$& $\tau'>b$\\
     \midrule
     $\cos(z(\cdot-a))$& $\left[1+\lambda r-\frac{(z^2-\lambda^2)r^2}{2}\right]e^{-\lambda r}$ & $\cos(z(\tau'-a))$& $\left[1+\lambda r-\frac{(z^2-\lambda^2)r^2}{2}\right]e^{-\lambda r}$ \\
     \addls
     $\sin(z(\cdot-a))$& $z(\tau'-a)(1+\lambda r) e^{-\lambda r}$& $\sin(z(\tau'-a))$& $z(\tau'-b)(1+\lambda r) e^{-\lambda r}$\\
     \bottomrule
    \end{tabular}
    \label{tab:hzx-52}
    \begin{tablenotes}
        \item[1] $r=\min\{|\tau'-a|,|\tau'-b|\}.$
    \end{tablenotes}
\end{threeparttable}
\end{table} 

\begin{table}[htbp]
    \centering
    \caption{\matern-$\tfrac{5}{2}$ VFRFs on Fourier basis $\phi(x;z)=\cos(z(x-a))$. }
    \setlength\tabcolsep{5pt}  
    \begin{tabular}{cl}
    \toprule
    $t'\in\RR$ & LFM Fourier Feature $P\circ k\circ G$ (cosine part)\\
    \midrule
    $t'<a$  
    & $ -\left[\frac{(z^2-\lambda^2)r_a^2}{2\beta(\gamma+\lambda)} + \frac{(z^2 - \gamma\lambda-2\lambda^2)r_a}{\beta(\gamma+\lambda)^2}
        + \frac{z^2 - \gamma^2 - 3\gamma\lambda-3\lambda^2}{\beta(\gamma+\lambda)^3}\right]e^{-\lambda r_a}$ \\    
    \addls
    $a\leq t'\le b$  
    & $\frac{\cos(z r_a+\theta)}{\beta\sqrt{z^2 + \gamma^2}}
        - \left[\frac{z^2 - \gamma^2 -3\gamma\lambda-3\lambda^2}{\beta(\gamma+\lambda)^3}+\frac{\gamma}{\beta(z^2+\gamma^2)}\right]e^{-\gamma r_a} $ \\
    \addls
    $t'>b\ (\gamma\neq\lambda)$ 
    & \makecell[l]{
        $- \left[\frac{z^2 - \gamma^2 -3\gamma\lambda-3\lambda^2}{\beta(\gamma+\lambda)^3}+\frac{\gamma}{\beta(z^2+\gamma^2)}\right]
        e^{-\gamma r_a}
         + \left[\frac{z^2 - \gamma^2 +3\gamma\lambda-3\lambda^2}{\beta(\gamma-\lambda)^3}+\frac{\gamma}{\beta(z^2+\gamma^2)}\right]
        e^{-\gamma r_b}$ \\
        \addlinespace[1pt]
        $ \quad\  -\left[\frac{(z^2-\lambda^2)r_b^2}{2\beta(\gamma-\lambda)}
        - \frac{(z^2+\gamma\lambda-2\lambda^2)r_b}{\beta(\gamma-\lambda)^2}
        + \frac{z^2-\gamma^2+3\gamma\lambda-3\lambda^2}{\beta(\gamma-\lambda)^3}\right]e^{-\lambda r_b}$
        }\\
    \addls
    $t'>b\ (\gamma=\lambda)$ 
    & $ -\left[\frac{z^2-7\lambda^2}{8\beta\lambda^3} +\frac{\lambda}{\beta(z^2+\lambda^2)}\right]e^{-\lambda r_a}
        + \left[\frac{\lambda}{\beta(z^2+\lambda^2)}-\frac{[(z^2-\lambda^2)r_b^2-3\lambda r_b-6]r_b}{6\beta}\right]e^{-\lambda r_b}$\\
    \bottomrule
    \end{tabular}
    \label{tab:mkg-52-cos}
\end{table}

\begin{table}[H]
    \centering
    \caption{\matern-$\tfrac{5}{2}$ VFRFs on Fourier basis $\phi(x;z)=\sin(z(x-a))$. }
    \begin{tabular}{cl}
    \toprule
    $t'\in\RR$ & LFM Fourier Feature $P\circ k\circ G$ (sine part)\\
    \midrule
    $t'<a$  
    & $ -\frac{z}{\beta}\left[\frac{\lambda r_a^2}{(\gamma+\lambda)}+\frac{(\gamma+3\lambda)r_a}{(\gamma+\lambda)^2}
        + \frac{\gamma+3\lambda}{(\gamma+\lambda)^3}\right]e^{-\lambda r_a}$ \\
    \addls
    $a\leq t'\le b$  
    & $\frac{\sin(z r_a+\theta)}{\beta\sqrt{z^2 + \gamma^2}}
        + \frac{z}{\beta}\left[\frac{1}{(z^2+\gamma^2)}-\frac{(\gamma+3\lambda)}{(\gamma+\lambda)^3}\right]e^{-\gamma r_a} $ \\
    \addls
    $t'>b\ (\gamma\neq\lambda)$ 
    & \makecell[l]{
        $ \frac{z}{\beta}\left[\frac{1}{(z^2+\gamma^2)}-\frac{(\gamma+3\lambda)}{(\gamma+\lambda)^3}\right]e^{-\gamma r_a}
         - \frac{z}{\beta}\left[\frac{1}{(z^2+\gamma^2)} - \frac{(\gamma-3\lambda)}{(\gamma-\lambda)^3}\right]e^{-\gamma r_b}$ \\
        \addlinespace[1pt]
        $\quad\ + \frac{z}{\beta}\left[\frac{\lambda r_b^2}
        {(\gamma-\lambda)}+\frac{(\gamma-3\lambda)r_b}{(\gamma-\lambda)^2}-\frac{\gamma-3\lambda}{(\gamma-\lambda)^3}\right]
        e^{-\lambda r_b}$
        }\\
    \addls
    $t'>b\ (\gamma=\lambda)$ 
    & $ \frac{z}{\beta}\left[\frac{1}{(z^2+\lambda^2)}-\frac{1}{2\lambda^2}\right]e^{-\lambda r_a}
        - \frac{z}{\beta}\left[\frac{1}{(z^2+\lambda^2)}-\frac{(2\lambda r_b+3)r_b^2}{6}\right]e^{-\lambda r_b}$\\
    \bottomrule
    \end{tabular}  
    \label{tab:mkg-52-sin}
\end{table}

\clearpage
\section{Experimental Details}
\label{app:experiment-details}

All models in the experimental section are implemented using GPyTorch \citep{gardner2018gpytorch}, trained by Adam Optimizer on an NVIDIA A100-SXM4 GPU (for TIMIT and UCI datasets) or Apple Macbook CPUs (for illustrative examples), with a learning rate of 0.01 and a batch size of 10,000. The models using doubly stochastic variational inference, e.g., IDDGPs, DLFM-VFRF, employ five samples for layer-wise sampling during training. We follow \cite{dgp2017} to set up a linear mean function for all the inner layers and a zero-mean function for the outer layer to avoid pathological behaviour \citep{duvenaud14avoid}. The weights of the linear mean function are fixed and determined by SVD if the input and output dimensions are not equal. The variational distributions over inducing variables are initialised to normal distributions with zero mean and variances identity for the outer layers and $10^{-5}$ for the inner layers. The inducing points are initialised with K-means. All models, including RFF-based models, used 100 Monte Carlo samples for evaluations on test data. 

Unless specifically stated, the RKHS interval is set to $[a,b]=[-1,4]$, and all input data are normalised to $[0,3]$. We initialise our model with length-scale $l=0.1$ for the TIMIT dataset and $l=1$ for the UCI datasets, ODE parameters $\alpha=1, \beta=0.01$, kernel variance $\sigma^2=0.1$ and noise variance $\varepsilon^2=0.01$.

\paragraph{Intermediate Outputs for Synthetic Data}  
We present here the posterior distributions of the DGP, the IDDGP and our model DLFM-VFRF on the synthetic data in Section~\ref{sec:EXP}.
\begin{figure}[h]
    \centering
    \includegraphics[width=0.75\linewidth]{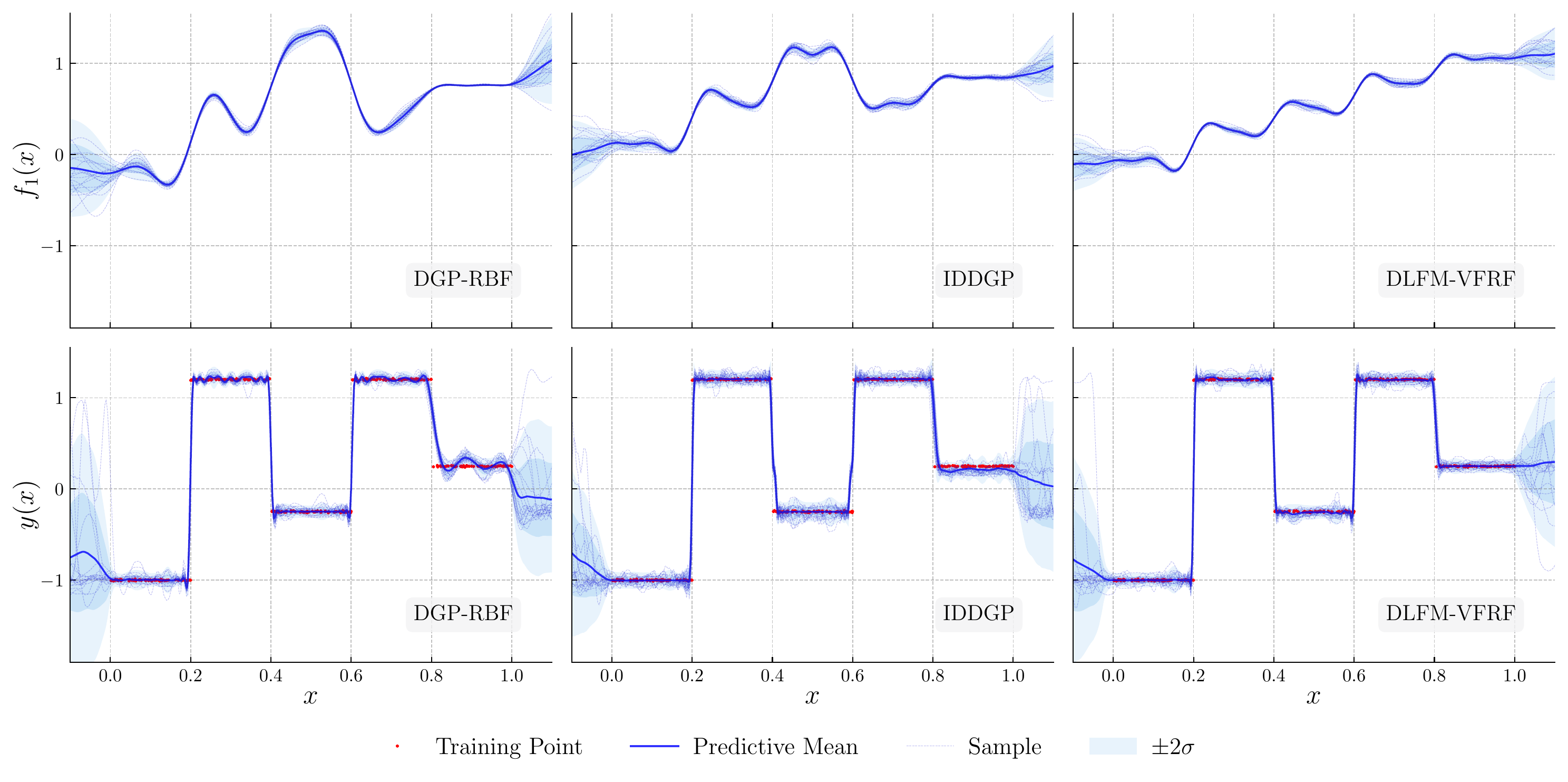}
    \caption{Comparison of posterior distributions of different compositional GP models on synthetic data. \textbf{Top row}: The output distributions of the intermediate layers. \textbf{Bottom row}: The posterior predictive distributions. Training points are marked with red dots, posterior means with blue lines, and uncertainty with varying shades of blue. Each panel depicts 20 samples from the posterior distribution. Although both the IDDGP and our model show better fitting to the multi-step function, they have very different intermediate posterior distributions.}
    \label{fig:hidden-layers}
\end{figure}

\clearpage

\end{document}